\documentclass[lettersize,journal]{IEEEtran}
\usepackage{amsmath,amsfonts}
\usepackage{algorithm}
\usepackage{array}
\usepackage[caption=false,font=normalsize,labelfont=sf,textfont=sf]{subfig}
\usepackage{textcomp}
\usepackage{stfloats}
\usepackage{url}
\usepackage{verbatim}
\usepackage{graphicx}
\usepackage{cite}
\usepackage{amssymb}

\usepackage{multirow}
\usepackage{adjustbox}
\usepackage{subfig}

\usepackage{algpseudocode}

\usepackage{rotating}
\usepackage{enumitem}

\usepackage{siunitx}
\usepackage{xcolor}
\definecolor{custom_green}{RGB}{0,170,0} 

\begin{document}

\title{NVS-SQA: Exploring Self-Supervised Quality Representation Learning for Neurally Synthesized Scenes without References}



\author{Qiang Qu, Yiran Shen*,~\IEEEmembership{Senior Member,~IEEE}, Xiaoming Chen*\thanks{* Corresponding authors}, Yuk Ying Chung, \\Weidong Cai, Tongliang Liu,~\IEEEmembership{Senior Member,~IEEE}
\IEEEcompsocitemizethanks{
\IEEEcompsocthanksitem Qiang Qu, Yuk Ying Chung, Weidong Cai, and Tongliang Liu are with School of Computer Science, the University of Sydney, Australia. E-mail: vincent.qu@sydney.edu.au; vera.chung@sydney.edu.au; tom.cai@sydney.edu.au; tongliang.liu@sydney.edu.au
\IEEEcompsocthanksitem Yiran Shen is with School of Software, Shandong University, China. 
E-mail: yiran.shen@sdu.edu.cn
\IEEEcompsocthanksitem Xiaoming Chen is with School of Computer and Artificial Intelligence, Beijing Technology and Business University, China.
E-mail: xiaoming.chen@btbu.edu.cn

}}

\markboth{Preprint}%
{Shell \MakeLowercase{\textit{et al.}}: A Sample Article Using IEEEtran.cls for IEEE Journals}


\maketitle

\begin{abstract}
Neural View Synthesis (NVS), such as NeRF and 3D Gaussian Splatting, effectively creates photorealistic scenes from sparse viewpoints, typically evaluated by quality assessment methods like PSNR, SSIM, and LPIPS. However, these full-reference methods, which compare synthesized views to reference views, may not fully capture the perceptual quality of neurally synthesized scenes (NSS), particularly due to the limited availability of dense reference views. Furthermore, the challenges in acquiring human perceptual labels hinder the creation of extensive labeled datasets, risking model overfitting and reduced generalizability. To address these issues, we propose NVS-SQA, a NSS quality assessment method to learn no-reference quality representations through self-supervision without reliance on human labels. Traditional self-supervised learning predominantly relies on the ``same instance, similar representation'' assumption and extensive datasets. However, given that these conditions do not apply in NSS quality assessment, we employ heuristic cues and quality scores as learning objectives, along with a specialized contrastive pair preparation process to improve the effectiveness and efficiency of learning. The results show that NVS-SQA outperforms 17 no-reference methods by a large margin (i.e., on average 109.5\% in SRCC, 98.6\% in PLCC, and 91.5\% in KRCC over the second best) and even exceeds 16 full-reference methods across all evaluation metrics (i.e., 22.9\% in SRCC, 19.1\% in PLCC, and 18.6\% in KRCC over the second best).
\end{abstract}


\section{Introduction} \label{sec:introduction}

\IEEEPARstart{P}{Photorealistic} view synthesis has emerged as a cornerstone in modern computer vision, bridging the gap between captured imagery and artificially rendered content for a wide array of applications such as film production, telepresence, robotics, and digital content creation~\cite{mildenhall2020nerf, han2019image, wang2024perf}. By generating novel viewpoints from existing data, recent Neural View Synthesis (NVS) techniques, including Neural Radiance Fields (NeRF)~\cite{mildenhall2020nerf} and 3D Gaussian Splatting~\cite{kerbl20233d}, have demonstrated remarkable promise in producing highly detailed and consistent scenes. This rapid progress underscores a growing imperative to rigorously evaluate the perceptual quality of neurally synthesized scenes (NSS), thereby guiding the development of robust NVS methods and advancing our understanding of how humans perceive visual content~\cite{liang2024perceptual, ma2018group, qu2024nerfnqa}.

\begin{figure}[htb]
  \centering
   \includegraphics[width=\linewidth]{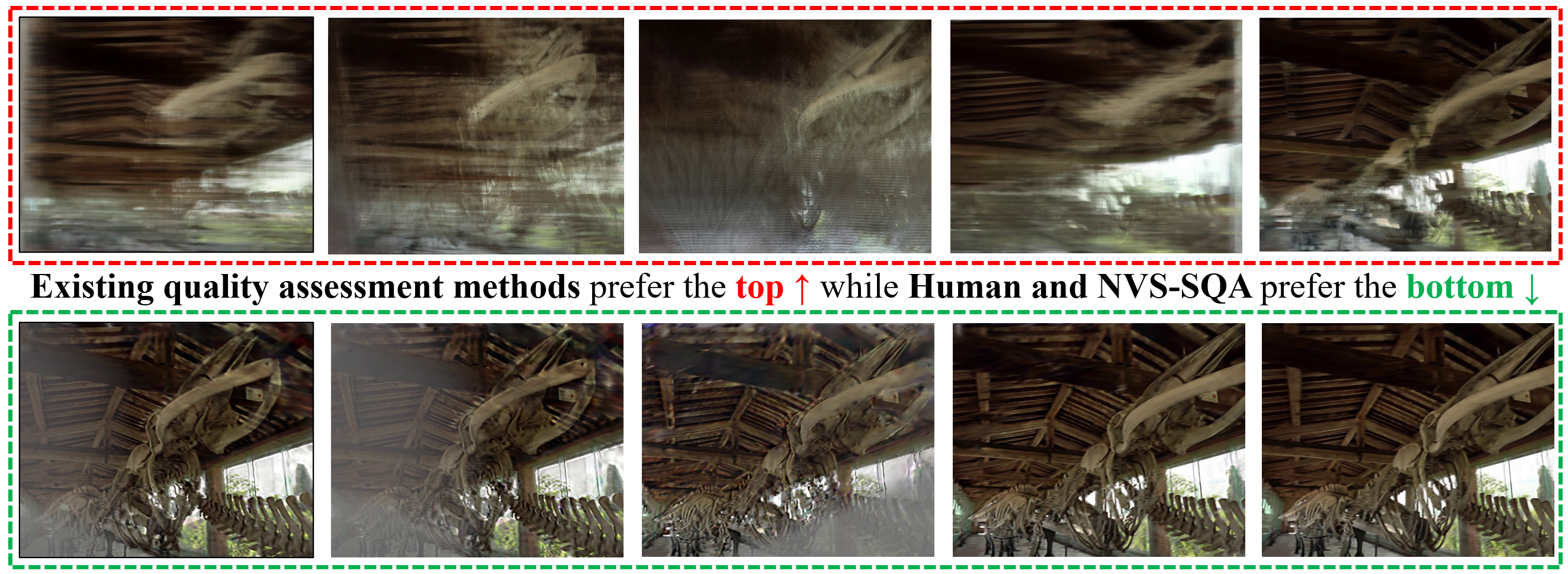}

   \caption{\textbf{Which synthesized scene is better (\textcolor{red}{top} or \textcolor{custom_green}{bottom})?} 
  The existing quality assessment methods display deviations from human judgment, including full-reference image (PSNR, SSIM, LPIPS) and video (VMAF, FovVideoVDP) quality assessment methods, no-reference image (BRISQUE, Re-IQA) and video (Video-BLIINDS, DOVER) quality assessment methods, and light-field quality assessment methods (ALAS-DADS, LFACon). Uniquely, the proposed method mirrors human subjective evaluations without references, trained with self-supervised learning.}

   \label{fig:teaser}
\end{figure}

\begin{figure}[htb]
  \centering
   \includegraphics[width=\linewidth]{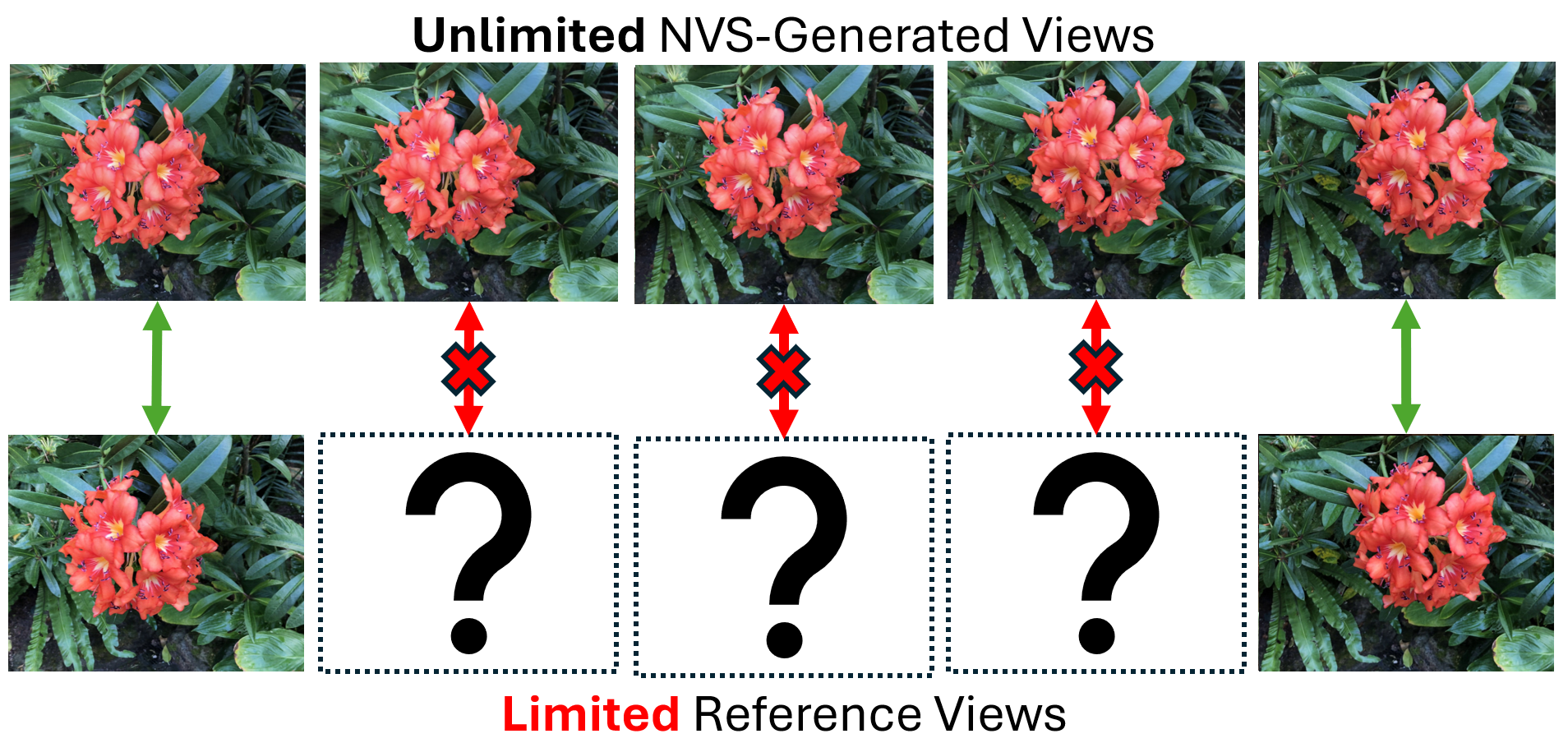}

   \caption{\textbf{Why no-reference?} Full-reference quality assessment methods (e.g., PSNR, SSIM) face a dilemma between the unlimited number of views in NSS and the limited availability of recorded references, whereas no-reference methods address this challenge.}

   \label{fig:why_no_reference}
\end{figure}

However, assessing the quality of NSS poses challenges, requiring comprehensive evaluations of spatial fidelity, view-to-view consistency, and perceptual quality~\cite{liang2024perceptual} (shown in Fig.~\ref{fig:teaser}). Current assessment approaches predominantly employ full-reference image quality methods, such as PSNR, SSIM \cite{wang2004image}, and LPIPS \cite{zhang2018unreasonable}. These methods require testers to designate a subset of views as reference images, against which NVS-generated views are compared, to evaluate the quality of a NSS. However, the scarcity of reference images for dense views further exacerbates the challenge of conducting a thorough quality assessment (as demonstrated in Fig.~\ref{fig:why_no_reference}). For example, datasets like LLFF~\cite{mildenhall2019local} and DTU~\cite{jensen2014large} provide only a limited number of reference views, inadequate for the evaluation of the densely synthesized views. Consequently, this limitation underscores the increasing necessity for no-reference methods in the quality assessment of NSS. Although no-reference methods are more relevant in practical scenarios, they face significant challenges, as they cannot directly evaluate the fidelity of the synthesized views against the references~\cite{mittal2012no, wu2023dover, wang2023exploring}.

Another challenge for designing NSS quality assessment methods is the limited availability of human perceptual annotations due to the extensive resources and time required for their collection, which leads to small labeled datasets~\cite{liang2024perceptual}. Such datasets elevate the risk of model overfitting and constrain the ability to generalize. For example, the Lab dataset~\cite{liang2024perceptual} includes only 70 final perceptual labels, an amount too modest to train deep learning models effectively. Finally, the variability in view counts and spatial resolutions within NSS complicates the development of an end-to-end deep learning model. All of those challenges prompt a question: \textit{is it possible to train a no-reference method for assessing the perceptual quality of NSS with minimal reliance on human annotation?}

\begin{figure}[htb]
  \centering
   \includegraphics[width=\linewidth]{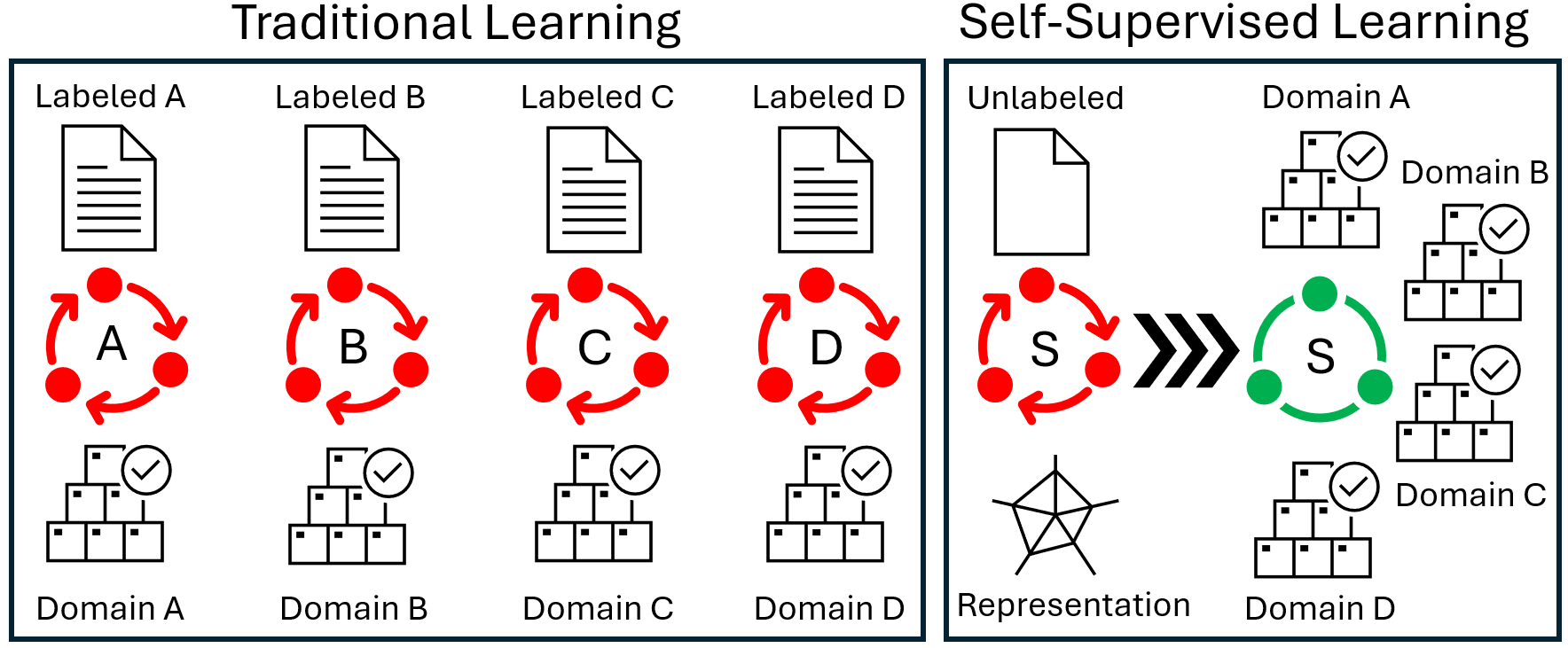}

   \caption{\textbf{Why self-supervised learning in NSS quality assessment?} Traditional learning-based quality assessment methods (left) require costly retraining for different domains (e.g., datasets, quality assessment protocols) and can easily overfit in the case of limited human perceptual labels. In contrast, self-supervised learning-based methods (right) learn generalized quality representations once from unlabeled data, allowing them to easily adapt to new domains without overfitting.}

   \label{fig:why_self_learning}
\end{figure}
To address these challenges, we introduce NVS-SQA, the first self-supervised learning framework for NSS quality assessment, offering distinct advantages as illustrated in Fig.~\ref{fig:why_self_learning}. This framework aims to derive effective no-reference quality representations for NSS from unlabeled datasets. NVS-SQA leverages heuristic cues and quality scores as soft learning objectives to calibrate the distance between the quality representations of contrastive pair (detailed in Section~\ref{subsec:multibranch_contrastive_objective}). To converges toward an optimal multi-branch guidance for quality representation learning, we design an adaptive weighting method inspired from the statistical principles. Furthermore, we develop a deep neural network as the backbone of our framework, designed to handle scenes of varying view counts and spatial resolutions. In the experiments, the proposed method was trained on an unlabeled dataset. Subsequently, its weights were frozen, and it was evaluated using linear regression on three labeled datasets comprising various new scenes produced by entirely unseen NVS methods (i.e., entirely distinct from the training dataset). The results demonstrate that the proposed approach surpasses 17 existing no-reference quality assessment methods in assessing NSS and even outperforms 16 mainstream full-reference quality assessments across six out of six evaluation metrics. The key points of this paper are summarized as follows:
\begin{itemize}[itemsep=2mm, parsep=0pt]
    \item We propose the first self-supervised learning framework for learning quality representations in NSS, demonstrating its capability to produce reference-free quality representations that generalize across different scenes and unseen NVS methods.

    \item Within the framework, we design a NSS-specific approach for preparing contrastive pairs, and multi-branch guidance adaptation for quality representation learning inspired by heuristic cues and full-reference quality scores to replace the inappropriate ``same instance, similar representation'' assumption.

    \item We introduce a benchmark for self-supervised learned quality assessment in NSS. We have open-sourced the project, including the source code, and datasets for self-supervised learning and evaluation, to facilitate future research in this field. Codes, models and demos are available at \url{https://github.com/VincentQQu/NVS-SQA}.

\end{itemize}

\section{Related work}
\label{sec:related_work}

Quality assessment in multimedia content has been a critical area of research, with methodologies broadly categorized into full-reference and no-reference approaches, depending on whether they require access to the original, unaltered media for comparison~\cite{ma2018group, qu2021light, qu2023lfacon}. Full-reference methods, as the name suggests, rely on a complete reference media for quality score prediction. Conversely, no-reference methods assess quality independently of the original media, a necessity in scenarios where the reference is unavailable or inapplicable~\cite{ma2018group, qu2023lfacon}. This distinction is particularly pertinent in Novel View Synthesis quality assessment, where datasets like LLFF present challenges for full-reference evaluations due to their sparse image availability~\cite{mildenhall2019local}. Our focus, therefore, is on advancing no-reference quality assessment techniques.

\noindent{\textbf{Image quality assessment.}}
The field of image quality assessment (IQA) has seen extensive exploration, yielding a variety of full-reference metrics tailored for 2D images. These include PSNR, SSIM~\cite{wang2004image}, MS-SSIM~\cite{wang2003multiscale}, IW-SSIM~\cite{wang2010information}, VIF~\cite{sheikh2006image}, FSIM~\cite{zhang2011fsim}, GMSD~\cite{xue2013gradient}, VSI~\cite{zhang2014vsi}, DSS~\cite{balanov2015image}, HaarPSI~\cite{reisenhofer2018haar}, MDSI~\cite{nafchi2016mean}, LPIPS~\cite{zhang2018unreasonable}, and DISTS~\cite{ding2020image}. These methods range from PSNR, which quantifies image reconstruction quality by comparing signal power against corrupting noise, to SSIM and its variants that evaluate perceptual aspects like structural integrity and texture. Advanced metrics like LPIPS leverage deep learning to capture nuanced visual discrepancies, offering insights into perceptual quality beyond traditional methods~\cite{zhang2018unreasonable}. For no-reference IQA, techniques such as BRISQUE~\cite{mittal2012no}, NIQE~\cite{mittal2012making}, PIQE~\cite{venkatanath2015blind} and CLIP-IQA~\cite{wang2023exploring} assess image quality by analyzing inherent image properties, with BRISQUE, for instance, evaluating naturalness degradation through local luminance statistics~\cite{mittal2012no}. The most recent no-reference quality assessment methods include CONTRIQUE~\cite{madhusudana2022image}, and Re-IQA~\cite{saha2023re}. CONTRIQUE\cite{madhusudana2022image} trains a deep Convolutional Neural Network (CNN) to learn robust and perceptually relevant representations, which a linear regressor then maps to quality scores in a no-reference setting. Similarly, Re-IQA introduces a Mixture of Experts approach, training two separate encoders to capture high-level content and low-level image quality representations achieving state-of-the-art performance on diverse image quality database~\cite{saha2023re}.

\begin{figure*}[b]
  \centering
   \includegraphics[width=\linewidth]{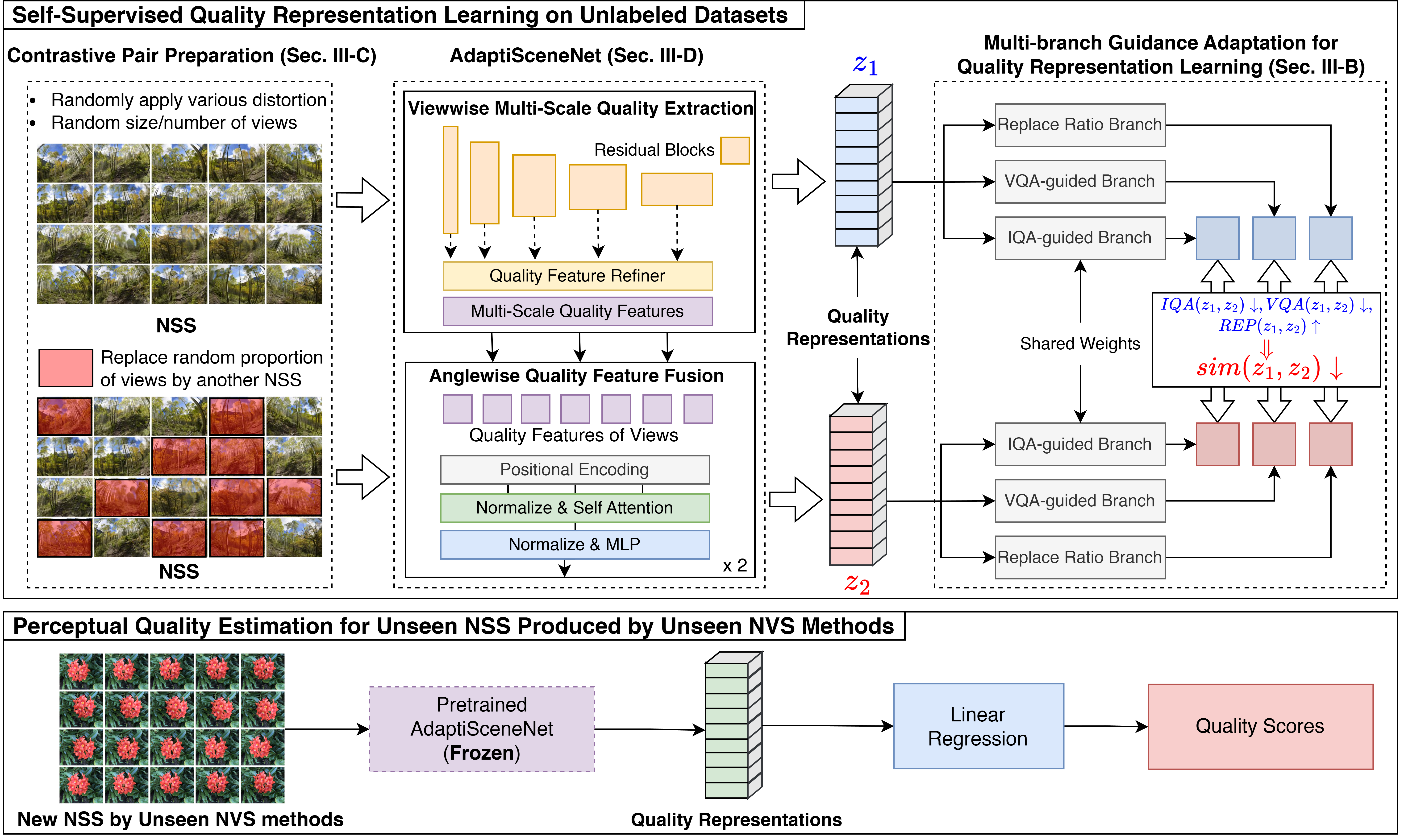}

   \caption{{\bf Overview of the proposed learning framework.} The framework consists of two main stages: (1) The self-supervised quality representation learning stage, where AdaptiSceneNet (described in Section~\ref{subsec:adaptiscenenet}) learns quality representations from unlabeled NSS through a contrastive pair preparation (detailed in Section~\ref{subsec:contrastive_pair_preparation}) process and multi-branch guidance adaptation for quality representation learning (described in Section~\ref{subsec:multibranch_contrastive_objective}), and (2) The perceptual quality estimation stage, where the pretrained AdaptiSceneNet is used to estimate perceptual quality by mapping learned representations to human scores.}

   \label{fig:framework}
\end{figure*}

\noindent{\textbf{Video quality assessment.}}
The assessment of video quality extends the principles of IQA to dynamic visual content, incorporating no-reference video quality assessment (VQA) methods such as Video-BLIINDS~\cite{saad2014blind}, VIIDEO~\cite{mittal2015completely}, FAST-VQA~\cite{wu2022fast}, FasterVQA~\cite{wu2023neighbourhood}, DOVER~\cite{wu2023dover}, DOVER-Mobile~\cite{wu2023dover} as well as full-reference methods such as STRRED~\cite{soundararajan2012video}, VMAF~\cite{li2016toward}, and FovVideoVDP~\cite{mantiuk2021fovvideovdp}. FAST-VQA introduces an efficient, end-to-end deep framework for VQA with a novel fragment sampling strategy, while FasterVQA further enhances computational efficiency and accuracy. DOVER disentangles the aesthetic and technical dimensions of user-generated content, offering an effective measure of perceived video quality. Meanwhile, VMAF fuses multiple metrics to align closely with human perception, and FovVideoVDP adopts a viewer-centric approach by accounting for gaze position and foveation. These techniques can be readily applied to NSS, treating synthesized view sequences in a manner analogous to video streams, thereby providing deeper insight into the perceptual quality of neurally rendered scenes.

\noindent{\textbf{Light-field quality assessment.}}
Light-field imaging (LFI) introduces an angular dimension to visual content, necessitating specialized assessment methods (LFIQA) such as NR-LFQA~\cite{shi2019no}, Tensor-NLFQ~\cite{zhou2020tensor}, ALAS-DADS~\cite{qu2021light} and LFACon~\cite{qu2023lfacon}. ALAS-DADS innovates with depthwise and anglewise separable convolutions for efficient and comprehensive quality assessment in immersive media~\cite{qu2021light}. LFACon further refines this approach by incorporating anglewise attention mechanisms, optimizing for both accuracy and computational efficiency in evaluating light-field image quality~\cite{qu2023lfacon}. These methodologies are particularly relevant for NSS, where synthesized views can be organized into a light-field matrix, mirroring the angular diversity of traditional light-field cameras.

\noindent{\textbf{NSS quality assessment.}}
The NSS quality assessment has predominantly relied on full-IQA methods, including PSNR, SSIM, and the perceptually-driven LPIPS~\cite{mildenhall2020nerf, barron2022mip, sun2022direct, fridovich2022plenoxels, gao2022mps, wizadwongsa2021nex, suhail2022light}. These methodologies facilitate a direct comparison between the synthesized images and their reference counterparts, serving as a metric for evaluating both similarity and perceptual quality. However, as highlighted in Section~\ref{sec:introduction}, such methods exhibit a bias towards the reserved reference views and fail to account for the dynamic quality inherent between views. In our previous work, we introduced NeRF-NQA\cite{qu2024nerfnqa}, the first fully supervised method for automatic NSS quality assessment. Although NeRF-NQA demonstrates promising performance, it depends heavily on human annotations gathered from relatively small datasets, raising concerns about overfitting and limited generalization. Moreover, its reliance on sparse points generated by COLMAP\cite{schonberger2016structure} undermines stability and precludes end-to-end training~\cite{qu2024nerfnqa}. To address these challenges, we propose NVS-SQA, the first self-supervised approach for NSS quality assessment. This study further evaluates how conventional IQA, VQA, and LFIQA methods compare to NeRF-NQA and NVS-SQA in effectively assessing the quality of neurally synthesized scenes.

\section{Methodology}
\label{sec:method}

\subsection{Problem Definition}
Let $S = \{s^1, s^2, ..., s^n\}$ be a neurally synthesized scene (NSS) composed of $n$ views generated by a Novel View Synthesis (NVS) method. The goal of Neurally Synthesized Scene Quality Assessment (NSS-QA) is to learn a function $f: S \rightarrow q \in \mathbb{R}$ where $q$ is a scalar score that reflects the perceptual quality of the scene as perceived by human observers. This score should account for comprehensive quality of NSS, including spatial fidelity, angular and temporal coherence, and structure integrity, capturing distortions that occur within individual views as well as inconsistencies across viewpoints. In the \textit{no-reference} setting, the function $f$ must estimate perceptual quality \textit{without access to any ground-truth or reference views}, relying solely on the synthesized scene itself. This makes the problem particularly challenging, as the model must infer perceptual degradation patterns directly from generated content. The predicted quality score is expected to satisfy three key properties: (1) strong correlation with human subjective ratings across varied scenes and synthesis methods, (2) sensitivity to perceptual artifacts including spatial degradations, temporal or angular inconsistencies, and geometric misalignments, and (3) generalization across scenes with diverse resolutions, view counts, and rendering pipelines. In this work, we address this problem through a self-supervised learning approach that leverages perceptually motivated guidance signals and contrastive representation learning to develop a robust no-reference quality estimator.

\subsection{Method overview}
\label{subsec:method_overview}

As illustrated in Fig.~\ref{fig:framework}, the proposed self-supervised, no-reference quality assessment methodology encompasses two primary stages: the self-supervised quality representation learning stage and the perceptual quality estimation stage.

The self-supervised quality representation learning stage aims to train a neural network capable of generating effective quality representations without the need for reference views or human perceptual annotations. During this stage, unlabeled NSS undergo a contrastive pair preparation process (elaborated in Section~\ref{subsec:contrastive_pair_preparation}) to create diverse pairs for subsequent training. These pairs are then feed to AdaptiSceneNet, a neural network designed for compatibility with varying scenes (detailed in Section~\ref{subsec:adaptiscenenet}), which processes the pairs separately but under shared weights to produce quality representation pairs. We introduce a multi-branch guidance adaptation (described in Section~\ref{subsec:multibranch_contrastive_objective}) to direct the quality representation learning process.

In the perceptual quality estimation stage (depicted on the bottom of Fig.~\ref{fig:framework}), the pretrained AdaptiSceneNet is employed on new NSS with its weights frozen. The perceptual quality estimation module then applies linear regression to map the quality representations output by AdaptiSceneNet to human perceptual scores. It is crucial to highlight that the quality representations are empirically validated to generalize across different scenes and NVS methods. This generalization capability is shown in the distinctiveness between the unlabeled dataset used for pretraining in the first stage and the dataset utilized for evaluation in the second stage, particularly in terms of scene variety and the NVS methods employed for scene generation.

\subsection{Multi-branch guidance adaptation for quality representation learning}
\label{subsec:multibranch_contrastive_objective}

\begin{figure}[htb]
  \centering
   \includegraphics[width=\linewidth]{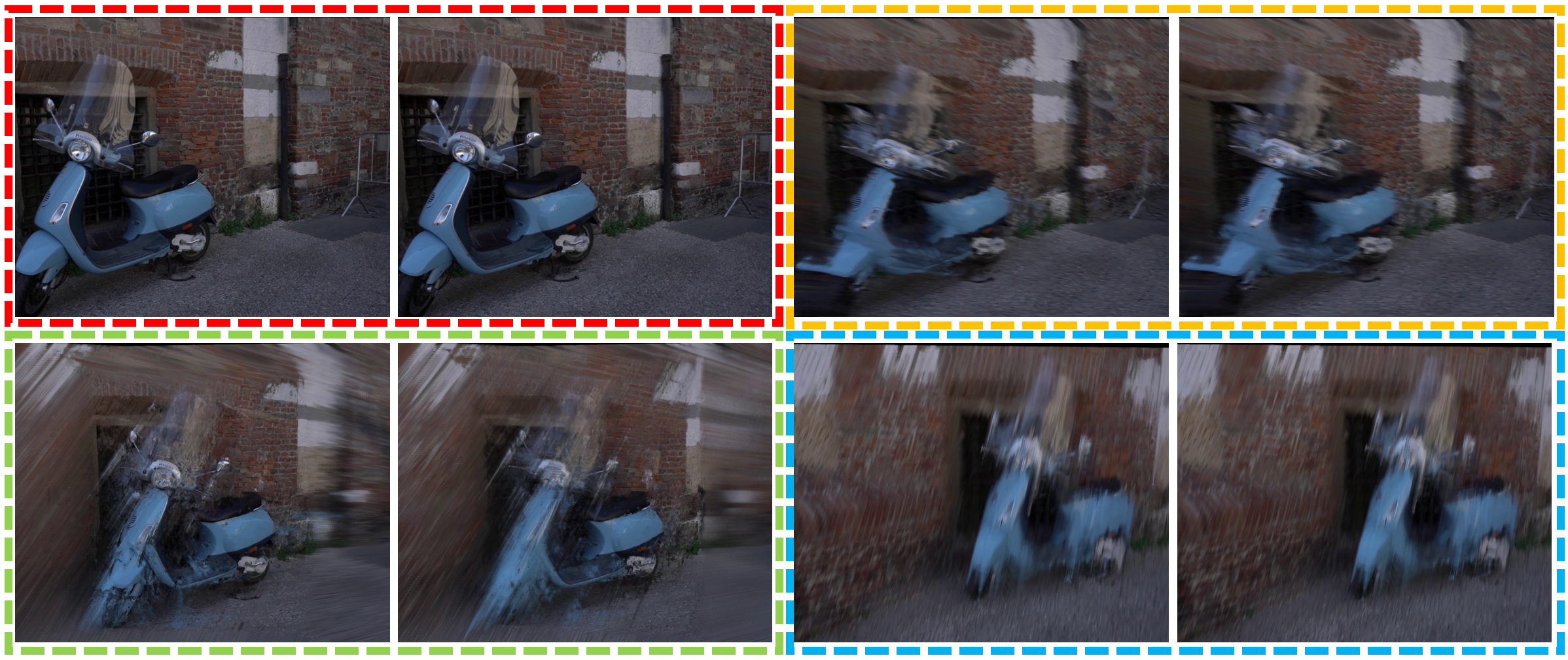}
   \caption{\textbf{Four clips from the same NSS but different quality level.} (zoom in for a clearer view)}
   \label{fig:same_nss_diff_quality}
\end{figure}

\begin{figure}[htb]
  \centering
   \includegraphics[width=\linewidth]{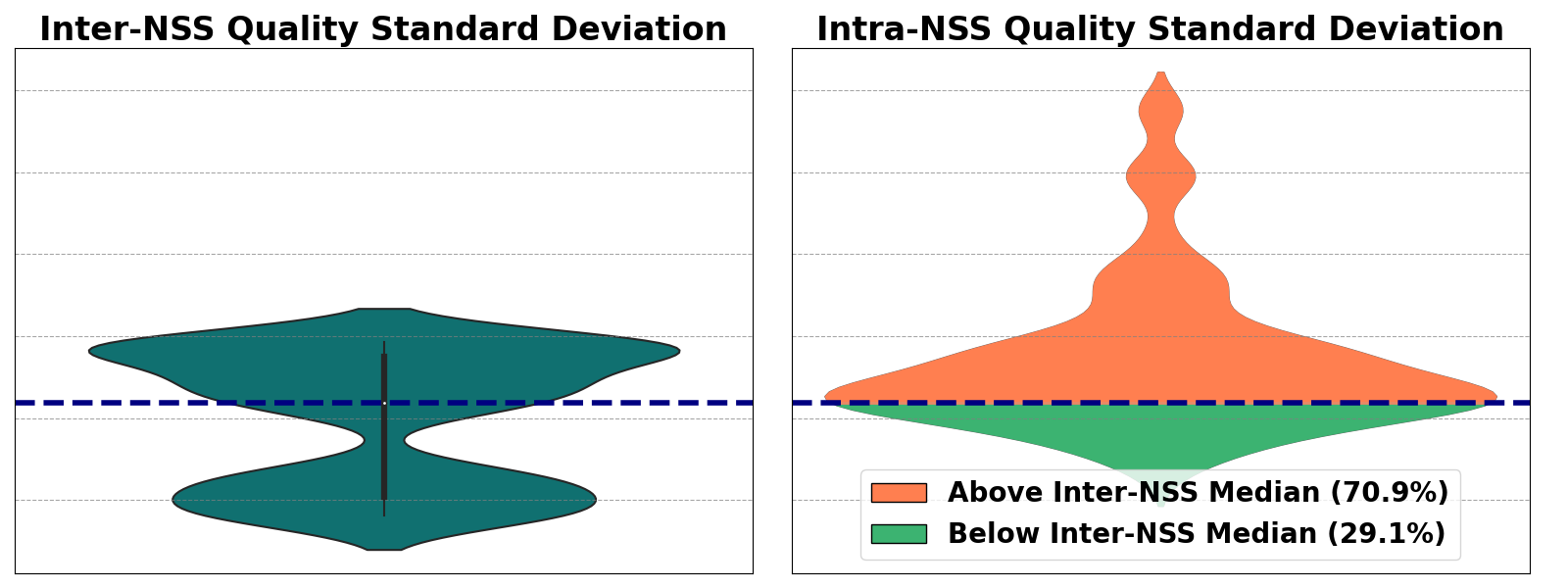}
   \caption{{\bf Intra-NSS quality variance can sometimes exceed inter-NSS variance.} This figure presents violin plots comparing the standard deviation of inter-NSS quality and intra-NSS quality within the LLFF dataset~\cite{mildenhall2019local}. The blue dashed line indicates the median of the intra-NSS quality standard deviation. The right figure shows that there are about 70\% of NSS exhibit greater internal quality variance than the median quality variance observed across different NSS.}

   \label{fig:inter_intra_std}
\end{figure}

Self-supervised learning needs appropriate objectives that enable the model to generate representations wherein the inputs sharing similar attributes yield similar representations~\cite{chen2020simple}. For instance, in semantic representation learning scenarios, crops derived from the same image are expected to possess similar representations~\cite{oord2018representation, chen2020simple, he2020momentum}. However, the assumption significantly diverges when applied to learning quality representations for NSS. It is not safe to presume that crops or clips from the same NSS instance share similar perceptual qualities. This is demonstrated by Fig. \ref{fig:same_nss_diff_quality} from the Fieldwork dataset~\cite{liang2024perceptual}, which shows that the quality of clips can vary significantly within the same NSS. One possible explanation is that the additional dimensions, compared to 2D images, introduce greater variability in quality across different clips. This is supported by the statistics of inter-NSS and intra-NSS quality shown in Fig.~\ref{fig:inter_intra_std}, where more than 70\% of NSS exhibit greater internal quality variance than the quality variance observed across different NSS in dataset LLFF~\cite{mildenhall2019local}. A deeper reason is that NVS methods are more susceptible to overfitting on the sparse training views provided in each scene~\cite{mildenhall2020nerf, barron2022mip, fridovich2022plenoxels, mueller2022instant}. This can lead to higher perceived quality in synthesized views that are closer to the training views. This observation raises a critical question: \textit{if the commonly held ``same instance, similar representation'' assumption is not valid, what can we rely on for self-supervised learning?} Another issue is that the effectiveness of self-learned representations hinges largely on the availability of a large corpus of images~\cite{chen2020simple, madhusudana2022image, saha2023re}. This requirement makes self-supervised learning more challenging for training with limited NSS datasets, where the volume of training examples may be insufficient to achieve generalized representation.

To address these issues, we design multiple  objectives for quality representation learning. The key idea is to utilize quality scores from various full-reference quality assessment methods, along with the proportion of replaced views in an NSS as complementary heuristic cues (detailed in Section~\ref{subsec:contrastive_pair_preparation}). Different from widely used discrete positive/negative labels for contrastive pairs~\cite{chen2020simple, he2020momentum, madhusudana2022image, saha2023re}, this approach provides continuous and comprehensive targets, enhancing the effectiveness of learning. Another advantage of this approach is that, unlike popular InfoNCE-like loss functions \cite{oord2018representation, chen2020simple, he2020momentum, madhusudana2022image, saha2023re}, it does not require loading a batch of negative pairs alongside a positive pair for each gradient update step (as shown in the equation~\ref{eq:contrastive_loss}), and hence reduce memory usage. This efficiency is particularly relevant in the context of the extensive data volumes associated with NSS.

The perceived quality of NSS is primarily influenced by three key factors: spatial fidelity, angular and temporal coherence, and structural integrity. To address these dimensions, we propose a multi-branch guidance adaptation strategy, illustrated in the right segment of Fig.~\ref{fig:framework}. In this design, the learned quality representations are passed through three distinct non-linear projectors, each guided by a different supervisory signal: an IQA-guided branch, a VQA-guided branch, and a replacement ratio (REP) branch. The IQA-guided branch leverages full-reference image quality assessment metrics to model spatial fidelity within individual views. The VQA-guided branch incorporates full-reference video quality assessment metrics to capture angular and temporal coherence across views. The REP branch provides a complementary cue based on the view replacement ratio, which reflects the proportion of synthesized views that are substituted due to occlusions or reconstruction defects. This ratio is typically indicative of structural degradation within the scene. Together, these branches provide complementary quality cues: the IQA branch emphasizes static spatial details, the VQA branch accounts for dynamic artifacts across views, and the REP branch captures structural inconsistencies and geometric distortions unique to multi-view synthesis. This multi-branch design enables the model to learn a robust and perceptually aligned embedding space.

To ensure notation consistency, we summarize the key symbols as follows:
\begin{itemize}
    \item \((s_1^i, s_2^i)\): the \(i\)-th contrastive pair of NSS samples;
    \item \(\Psi := \{\text{IQA}, \text{VQA}, \text{REP}\}\): the set of guidance branches;
    \item \(H\): the shared representation network;
    \item \(\tau_{\psi}\): branch-specific MLP projection for \(\psi\);
    \item \(\psi(s_1^i, s_2^i)\): the rescaled guidance signal (in \([-1, 1]\)) for branch \(\psi\) on pair \((s_1^i, s_2^i)\);
    \item \(\text{sim}(u, v)\): cosine similarity, defined as \(\frac{u^\top v}{\|u\|\|v\|}\);
\end{itemize}

Formally, given a set of contrastive pairs, \(\{ (s_1^i, s_2^i) \}_{i=1}^N\), and a representation model \(H\), we define each guidance branch \(\psi \in \Psi\), with branch-specific projector \(\tau_{\psi}\), and define the learning loss for each pair as:

\begin{equation}
\label{eq:contrastive_loss}
\begin{aligned}
\ell_{\psi}^i (H) = \left| \; \text{sim}\left( \tau_{\psi} H(s_1^i), \; \tau_{\psi} H(s_2^i) \right) \; - \; \psi(s_1^i, s_2^i) \; \right|,
\end{aligned}
\end{equation}

\noindent where \(\text{sim}(u, v) = \frac{u^\top v}{\|u\|\|v\|}\) denotes the cosine similarity between \(l_2\)-normalized vectors \(u\) and \(v\). The outputs of the IQA, VQA, and REP branches are all linearly rescaled to the range $[-1, 1]$ to match the similarity scale. Specifically, for a given scalar quality score $x$ in the range $[a,b]$, the rescaled value $\tilde{x}$ is computed as $\tilde{x} = 2 \cdot \frac{x - a}{b - a} - 1$. The values of $a$ and $b$ are defined empirically based on the score distribution in the training set for each branch.

It is important to note that the three guidance cues serve as soft and complementary supervision signals to guide the learning of quality-aware representations. The purpose is not to replicate the behavior of any individual cue but to leverage their distinct perspectives to shape a representation space that aligns more closely with human perceptual judgments. Each of the three guidance branches corresponds to a distinct aspect of perceptual quality. The IQA-guided branch captures spatial fidelity within individual synthesized views, where artifacts such as blurring or texture inconsistencies commonly occur. The VQA-guided branch emphasizes angular and temporal coherence across viewpoints, which helps detect issues like flickering, reprojection errors, or perceptual instability that arise from view-to-view inconsistencies. The REP-guided branch focuses on structural integrity by reflecting disruptions introduced through controlled view substitutions, such as geometric misalignments or parallax distortions. These three guidance branches act as soft supervision signals rather than strict labels, allowing the model to align the representational distance with perceptual degradations across a range of synthesis artifacts. By jointly leveraging all the three branches, NVS-SQA learns a more comprehensive and perceptually robust quality-aware embedding space, which is further supported by the quantitative and qualitative improvements observed in the ablation study (Section~\ref{subsec:ablation_study}).

While our approach adopts cosine similarity as a contrastive objective, it diverges significantly from prior contrastive learning methods such as SimCLR~\cite{chen2020simple}. SimCLR is designed for semantic representation learning from individual images, where the assumption is that different augmentations of the same image share the same semantics. In contrast, our approach operates on multi-view NSS, where the goal is not to preserve semantics but to learn perceptual quality embeddings that reflect not only intra-view quality but also view-dependent distortions, such as geometric inconsistencies, occlusions, and synthesis artifacts. Cosine similarity is retained in our formulation not to mimic SimCLR, but because it provides a scale-invariant, bounded similarity metric that remains effective under the variability of view count and spatial resolution in novel view synthesis. In our context, cosine distance serves as a perceptual proxy between multi-view NSS clips, enabling the model to learn quality-aware representations that are sensitive to structural degradations and cross-view inconsistencies.

\noindent{\textbf{Manual Branch Weighting (MBW).}}~\label{subsubsec:mbw}
An intuitive approach to learning in a multi-branch scenario involves manually assigning weights and experimenting with different weight combinations to find the optimal configuration. The objective is thus to learn an optimal representation model \(H^*\) such that:

\begin{equation}
\label{eq:obj}
\begin{aligned}
H^* = \arg \min_{H} \frac{1}{N} \sum_{i=1}^{N} \sum_{\psi \in \Psi} \left ( \lambda_{\psi} \ell_{\psi}^i (H) \right ),
\end{aligned}
\end{equation}

\noindent where \(\lambda_{ \psi}\) denotes the weight assigned to each branch-specific loss component, which can be adjusted via grid search (see ablation in Section~\ref{subsec:ablation_study}). This formulation adaptively combines heuristic guidance cues to optimize the quality representation.

\noindent{\textbf{Adaptive Quality Branching (AQB).}}~\label{subsubsec:aqb}
The MBW approach requires expensive training for each weight combination and makes it challenging to find the optimal configuration within a limited discrete search space. To address this, we adopt an automatic weighting method, AQB, inspired by \cite{kendall2017uncertainties}, that converges toward an optimal configuration efficiently. The likelihood for each branch, \(\psi \in \Psi\), can be defined using a Gaussian distribution:

\begin{equation}
\label{eq:single_dist}
\begin{aligned}
p(\psi(s_1^i, s_2^i) | \delta_{H, \psi}(s_1^i, s_2^i)) = \mathcal{N} \left( \delta_{H, \psi}(s_1^i, s_2^i), \sigma_{\psi}^2 \right),
\end{aligned}
\end{equation}
where \( \delta_{H, \psi}(s_1^i, s_2^i) := \text{sim} \left( \tau_{\psi} H(s_1^i), \; \tau_{\psi} H(s_2^i) \right) \) represents the predicted similarity for branch \(\psi\), and \(\sigma_{\psi}\) is the learnable noise parameter for branch \(\psi\).

The joint multi-branch probabilistic model is then:

\begin{equation}
\label{eq:joint_dist}
\begin{aligned}
p( \{\psi(s_1^i, s_2^i)\}_{\psi \in \Psi} | \{\delta_{H, \psi}(s_1^i, s_2^i)\}) = \prod_{\psi \in \Psi} {\mathcal{N} \left( \delta_{H, \psi}(s_1^i, s_2^i), \sigma_{\psi}^2 \right) }.
\end{aligned}
\end{equation}

\noindent The log-likelihood for the joint multi-branch probabilistic model~\cite{young2005essentials} can then be written as:
\begin{equation}
\label{eq:log_like}
\begin{aligned}
&\log p(\{\psi(s_1^i, s_2^i)\}_{\psi \in \Psi} | \{\delta_{H, \psi}(s_1^i, s_2^i)\}) \propto \\ &-\sum_{\psi \in \Psi} \left[ \frac{1}{2\sigma_\psi^2} \left( \psi(s_1^i, s_2^i) - \delta_{H, \psi}(s_1^i, s_2^i) \right)^2 + \log \sigma_\psi \right ].
\end{aligned}
\end{equation}

The learning objective is then minimizing the negative log-likelihood over all pairs:

\begin{equation}
\label{eq:obj_noise}
\begin{aligned}
H^* = \arg \min_{H, \{\sigma_\psi\}} \frac{1}{N} \sum_{i=1}^{N} \sum_{\psi \in \Psi} \Tilde{\ell}_{\psi}^i(H, \sigma_\psi) ,
\end{aligned}
\end{equation}
where the loss function for each branch is defined as:

\begin{equation}
\label{eq:contrastive_loss_noise}
\begin{aligned}
\Tilde{\ell}_{\psi}^i(H, \sigma_\psi) = \frac{1}{2\sigma_{\psi}^2} \left| \; \text{sim} \left( \tau_{\psi} H(s_1^i), \; \tau_{\psi} H(s_2^i) \right) - \psi(s_1^i, s_2^i) \right|^2 \\+ \log \sigma_{\psi},
\end{aligned}
\end{equation}

\noindent where each \(\sigma_{\psi}\) dynamically scales its respective loss branch and regularizes the overall loss, ensuring balanced training across IQA, VQA, and REP.

By integrating branch-wise noise into the loss function, we ensure that the model adaptively balances the learning process across different branches, reducing the dominance of any single branch and enabling optimal performance across all branches. The refined objective function properly weights each loss component by its corresponding noise, leading to a balanced and robust training process that leverages the shared quality representations more effectively. The optimal results for both MBW and AQB are presented in the experimental results in Section~\ref{subsec:ablation_study}.

\subsection{Contrastive pair preparation}
\label{subsec:contrastive_pair_preparation}

The preparation of pairs is crucial for acquiring robust representations, particularly in semantic representation learning. For instance, SimCLR \cite{chen2020simple} utilizes data augmentation techniques such as random cropping, color distortion, and Gaussian blur, designating crops from the same image instance as positive pairs and others as negative. However, these methods may not be sufficient within the context of NSS. Given the limited availability of NSS, the likelihood of generalizing self-supervised learned representations is reduced, highlighting the need for a more varied set of contrastive pairs from a restricted dataset. As discussed in Section~\ref{subsec:multibranch_contrastive_objective} and illustrated in Fig.~\ref{fig:same_nss_diff_quality}, the ``same instance, similar representation'' assumption proves unreliable for quality representation learning in NSS, making traditional positive/negative partitioning inappropriate.

To address these challenges, we have developed an algorithm for the preparation of contrastive pairs specifically tailored to NSS, as depicted in the leftmost block of Fig.~\ref{fig:framework}. According to the objectives outlined in Section~\ref{subsec:multibranch_contrastive_objective}, we do not classify pairs as positive or negative, allowing us to create NSS pairs at any quality distance and maximize diversity by applying varying levels of distortion. We further randomize crops and clips to ensure the model accommodates NSS with different numbers of views and spatial resolutions. Additionally, we randomly replace views with those from another NVS method to enhance variability. Importantly, to facilitate the effectiveness of quality representation learning, we ensure that each NSS pair represents the same scene, thereby controlling for semantic consistency.

Formally, the proposed pair preparation algorithm is outlined in Algorithm~\ref{alg:contrastive_pairs} titled \textit{Preparing Contrastive Pairs}. Each pair, maintaining identical semantic information, captures variations in scene quality through controlled distortions and modifications. Starting with an unlabeled dataset \(D\) and a target of \(N\) pairs, the algorithm creates a set \(P\) of pairs. It selects an NSS randomly from \(D\), applies random cropping and orientation adjustments to form a base scene \(s_1\), and then duplicates, distorts, and partially replaces views in \(s_1\) with those generated by an alternate NVS method to produce its counterpart \(s_2\). These steps introduce necessary variability and contrast, essential for training models to discern subtle visual differences. The algorithm also adjusts the number of views and spatial resolutions randomly, ensuring the model's compatibility with various NSS sizes. The replacement ratio, $r$, moderates the disparity between \(s_1\) and \(s_2\), serving as a heuristic cue for self-supervised learning (as detailed in Section~\ref{subsec:multibranch_contrastive_objective}). In our implementation, the replacement ratio $r$ is randomly sampled from a uniform distribution $r \sim \mathcal{U}(0, 0.5)$. This ratio controls the percentage of views in a base scene $s_1$ that are replaced with views synthesized by a different NVS method to create $s_2$. Through empirical exploration, the range $[0, 0.5]$ allows for diverse and controllable quality differences while avoiding redundancy, since replacement ratios beyond $0.5$ are equivalent to swapping the roles of $s_1$ and $s_2$. All replacements are performed within the same scene to maintain semantic consistency, and the replacement methods are selected randomly to ensure balance and reduce bias. By repeatedly forming such pairs, the algorithm compiles a diverse set \(P\) ready for subsequent self-supervised learning. Moreover, there are several designs to prevent unintended biases from disparities across NVS methods. For example, all view replacements are restricted to the same scene to ensure content alignment. The substituting NVS methods are selected uniformly at random during training to avoid overexposure to any single method. Additionally, the model is not supervised to identify or rank specific methods. Instead, soft quality cues, including image-level (IQA), video-level (VQA), and structural (REP) signals, are used to guide representation learning. These cues enable the model to learn perceptual quality embeddings that generalize beyond individual method characteristics.

To analyze the scalability of the proposed contrastive pair preparation, we consider a training setup with $s$ scenes, each containing $v$ distinct view sequences (i.e., consistent camera paths or spatial neighborhoods). For each view sequence, $m$ different NVS methods generate corresponding multi-view clips. From each NVS rendering, $c$ clips are extracted through randomized sampling and pre-processing. This results in $mc$ clips per view sequence. Contrastive pairs are only formed between clips corresponding to the same view sequence, ensuring semantic alignment while allowing variation across NVS methods and augmentations. The number of unordered clip pairs per view sequence is $P_{\text{view}} = \binom{mc}{2} = [mc (mc - 1)]/2$. The total number of base contrastive pairs across all view sequences and scenes is 
$P_{\text{total}} = s \cdot v \cdot [mc (mc - 1)]/2$. Each contrastive pair can then be further diversified using $a$ augmentation variants, including geometric transformations, appearance distortions, resolution changes, and view-level replacements. The effective number of unique training signals becomes $P_{\text{effective}} = a \cdot s \cdot v \cdot [mc (mc - 1)]/2$. For example, using conservative values such as $s = 10$ scenes, $v = 300$ view sequences per scene, $m = 5$ NVS methods, $c = 5$ clips per method, and $a = 20$ augmentation variants, the total effective training signal reaches tens of millions of unique contrastive pairs. This demonstrates that even under a limited number of scenes, the proposed pair preparation strategy enables large-scale self-supervised learning through combinatorial growth.

\begin{algorithm}
\small
\caption{\textit{Preparing Contrastive Pairs}}
\label{alg:contrastive_pairs}
\begin{algorithmic}[1]
\Require Unlabeled dataset \(D\) of NSS, number of result pairs \(N\).
\Ensure Set of contrastive pairs \(P\).

\State Initialize \(P \gets \emptyset\)

\For{$i \gets 1$ to $N$}
    \State $nss \gets \text{RandomlySelect}(D)$
    \State $views \gets \text{RandomlySelectViews}(nss)$
    \State $s_1 \gets \text{RandomCrop}(views)$
    \State $s_1 \gets \text{RandomRotateOrFlip}(s_1)$
    \State $s_2 \gets \text{ReplaceViews}(s_1)$
    \State $s_1, \, s_2 \gets \text{RandomDistortOrNo}(s_1), \, \text{RandomDistortOrNo}(s_2)$
    \State $s_2 \gets \text{ReplaceViews}(s_2, \text{AnotherRandomNVS}(views), r)$
    \State $P \gets P \cup \{(s_1, s_2)\}$
\EndFor

\State \Return $P$
\end{algorithmic}
\end{algorithm}


\subsection{AdaptiSceneNet}
\label{subsec:adaptiscenenet}
As mentioned in the previous sections, NSS encompass a wide range of views with varying spatial resolutions, requiring a versatile model architecture to accommodate these differences effectively. To this end, we developed AdaptiSceneNet, which integrates spatial convolutions with Transformer encoders~\cite{vaswani2017attention}, as depicted in Fig.~\ref{fig:framework}. AdaptiSceneNet comprises two primary modules: the Viewwise Multi-Scale Quality Extraction module and the Anglewise Quality Feature Fusion module. The Viewwise Multi-Scale Quality Extraction module utilizes several residual convolutional blocks~\cite{he2016deep}, refining output features through additional convolutions to capture multi-scale quality features from each view, from low to high-level attributes. This methodical approach to feature extraction, inspired by LPIPS~\cite{zhang2018unreasonable}, ensures a thorough quality analysis across scales and includes an adaptive average pooling layer to handle varying resolutions~\cite{goodfellow2016deep}. The Anglewise Quality Feature Fusion module employs positional encoding and several layers of transformer encoding, to effectively integrate these quality features within the angular domain.

\subsection{Self-supervised training procedure}

To complement the contrastive pair generation described in Algorithm \ref{alg:contrastive_pairs}, we present the self-supervised training procedure in Algorithm \ref{alg:training} for NVS-SQA, which includes batch-wise pair generation, multi-branch loss computation, and model updates using the AdaptiSceneNet backbone.

\begin{algorithm}
\small
\caption{\textit{Self-Supervised Training Procedure}}
\label{alg:training}
\begin{algorithmic}[1]
\Require Unlabeled dataset \(D\) of NSS, number of training epochs \(E\), batch size \(B\), learning rate \(\eta\)
\Ensure Trained quality representation model \(H\)

\State Initialize model weights for AdaptiSceneNet

\For{$\text{epoch} \gets 1$ to $E$}
    \State $P \gets \text{GenerateContrastivePairs}(D, \text{batch size} = B)$ \Comment{Using Algorithm~\ref{alg:contrastive_pairs}}
    \For{each $(s_1, s_2) \in P$}
        \State Compute quality representations $H(s_1), H(s_2)$
        \For{each branch} $\psi \in \{\text{IQA}, \text{VQA}, \text{REP}\}$
            \State Compute branch loss $\ell_\psi$ using Eq.~\ref{eq:contrastive_loss_noise}
        \EndFor
        \State Aggregate total loss
        \State Backpropagate and update network weights
    \EndFor
\EndFor

\State \Return $H$
\end{algorithmic}
\end{algorithm}

\noindent{\textbf{Time complexity analysis.}}
Let $N$ be the number of training scenes, $V$ the number of views per scene, $R \times C$ the spatial resolution of each view, $d$ the feature dimension, and $L$ the number of transformer layers. The contrastive pair generation step (Algorithm \ref{alg:contrastive_pairs}) has complexity $\mathcal{O}(N)$, as each scene is processed independently. For each training step, the AdaptiSceneNet backbone first applies viewwise convolutions with complexity $\mathcal{O}(B \cdot V \cdot R \cdot C)$ for a batch size $B$. It then performs transformer-based angular fusion with self-attention, which has complexity $\mathcal{O}(B \cdot L \cdot V^2 \cdot d)$. MLP heads have negligible cost in comparison. Hence, the total per-batch time complexity is $\mathcal{O}(B \cdot V \cdot R \cdot C + B \cdot L \cdot V^2 \cdot d)$. Despite the quadratic scaling in $V$, the model training remains computationally practical. On an RTX 4090 GPU, the average training time per NSS pair is approximately 8 seconds. During inference, our method requires only 0.692 seconds per scene on average. Further details about computational cost analysis are provided in Section \ref{subsec:computational_cost}.

This study involved human subjects for the collection of perceptual ratings in the newly introduced dataset (as described in Section~\ref{subsec:robustness_test}). All experimental procedures and protocols were approved by the Research Ethics Review Committee of BTBU. Informed consent was obtained from all participants prior to data collection.

\section{Experiments and results}
\label{sec:experiment}

\subsection{Datasets and evaluation protocol}

\noindent{\textbf{Unlabeled dataset for self-supervised learning.}}
Our model was trained on a compact dataset derived from Nerfstudio~\cite{nerfstudio}, encompassing 17 scenes, each synthesized using eight NVS methods and the respective variants. These included TensoRF~\cite{chen2022tensorf}, Instant-NGP~\cite{mueller2022instant}, K-Planes~\cite{fridovich2023k} (with the far value set to 20, 100, 1000), Nerfacto-default~\cite{nerfstudio}, Nerfacto-huge~\cite{nerfstudio}, and 3D Gaussian Splatting~\cite{kerbl20233d}. This selection was strategically made to encompass NVS methods that diverge significantly from those employed within the evaluation dataset, thereby examining the proposed method's capability to generalize across different NVS methods. The scenes within this dataset vary considerably in their complexity, with the number of views per scene ranging from 120 to 600. This diversity in scene composition and viewpoint count serves to test the adaptability and generalization of our self-supervised learning model across a broad spectrum of NVS-generated content.

\noindent{\textbf{Training data diversity and pair generation scale.}}
In addition to this scalable supervision, the scenes themselves were carefully selected to maximize real-world diversity. Specifically, these scenes were obtained using different devices such as mobile phones and mirrorless cameras with fisheye lenses~\cite{nerfstudio}. Importantly, this dataset is not limited to canonical forward-facing captures and includes 360-degree scenes with varying levels of quality and structural complexity. Unlike other datasets that focus on a central object or object-centric environments, the scenes span both indoor and outdoor settings, sparse and cluttered layouts, and complex motion paths~\cite{nerfstudio}. This ensures that our training data reflect the diversity and imperfections encountered in real-world applications. Finally, the high variability in view counts, spatial resolution, scene structure, and camera trajectories further contributes to the diversity of training signals. These factors, combined with our contrastive pair generation strategy, allow the model to generalize beyond the training distribution, as demonstrated by its consistent performance on multiple unseen datasets.

\noindent{\textbf{Labeled datasets for evaluation.}}
The evaluation of NVS-SQA leveraged three distinct labeled NSS datasets—Lab~\cite{liang2024perceptual}, LLFF~\cite{mildenhall2019local}, and Fieldwork~\cite{liang2024perceptual}—each presenting unique challenges. The Lab dataset features six real scenes captured with a 2D gantry system in a laboratory setting, offering a uniform grid of training views and reference videos ranging from 300 to 500 frames for a thorough quality assessment. In contrast, the LLFF dataset comprises eight real scenes captured with a handheld cellphone, providing a sparse selection of 20-30 test views per scene, with positional data computed via COLMAP~\cite{schonberger2016structure}. The Fieldwork dataset includes nine real scenes from diverse environments, such as outdoor urban landscapes and indoor museum settings, characterized by intricate backgrounds and variable lighting conditions, with reference videos typically containing around 120 frames. To synthesize NSS for each scene, ten NVS methods were utilized, spanning a variety of models with explicit and implicit geometric representations, rendering models, and optimization strategies, ensuring an unbiased evaluation of the framework's generalizability. Moreover, methods such as NeRF~\cite{mildenhall2020nerf}, Mip-NeRF~\cite{barron2022mip}, DVGO~\cite{sun2022direct}, Plenoxels~\cite{fridovich2022plenoxels}, NeX~\cite{wizadwongsa2021nex}, LFNR~\cite{suhail2022light}, IBRNet~\cite{wang2021ibrnet}, and GNT~\cite{wang2022attention} were involved, including both cross-scene (GNT-C and IBRNet-C) and scene-specific (GNT-S and IBRNet-S) models.
The human perceptual labels were collected in a comparison-based manner and in format of Just-Objectionable-Difference (JOD) units~\cite{liang2024perceptual}. These JOD scores were obtained from publicly available datasets~\cite{liang2024perceptual}, which were constructed through controlled user studies. The labeling process involved pairwise comparisons of novel-view video sequences, with the resulting perceptual scores scaled into JOD units using Thurstone’s Case V psychometric model. This methodology produced continuous, scene-level perceptual quality estimates  (Refer to~\cite{liang2024perceptual} for a full description of the experimental design, scaling methodology, and subject demographics). It is important to highlight that the JOD scale's utility and the comparative analysis it facilitates are particularly pronounced when conducted within the context of individual scenes~\cite{liang2024perceptual}.

\noindent{\textbf{Evaluation protocol.}}
We largely adhere to the evaluation protocol from \cite{madhusudana2022image, saha2023re, qu2023lfacon}, but in a more stringent manner. Our evaluation consists of two primary objectives: first, to determine if the proposed method can simultaneously adapt to three different datasets; second, to test the method's ability to perform cross-dataset validation. Following the self-supervised learning phase, the model's weights are frozen, and the model is applied to the three evaluation datasets to generate quality representations. A one-time linear regression is performed on a randomly sampled half of the integrated dataset to align these quality representations with human scores. The model's performance is then evaluated on the remaining half of the dataset to assess the generalizability of the NVS-SQA framework across the integrated datasets. This approach is designed to thoroughly evaluate the framework's capability to adapt and perform effectively amidst the varied complexities posed by the Lab, LLFF, and Fieldwork datasets simultaneously.  Additionally, we conduct cross-dataset experiments to further assess the adaptability of the proposed method across various datasets. Importantly, the Lab and Fieldwork datasets were created and labeled by Liang et al. ~\cite{liang2024perceptual} and are publicly available. The LLFF dataset constructed by Mildenhall et al.~\cite{mildenhall2019local} is also publicly accessible. All three datasets were constructed by independent research groups prior to our work. This ensures that the cross-dataset evaluation reflects realistic generalization to external and independently produced NSS data.

\noindent{\textbf{Evaluation metrics.}}
As two of the most widely used metrics in quality assessment~\cite{madhusudana2022image, saha2023re, wu2022fast}, the Spearman Rank Order Correlation Coefficient (SRCC)~\cite{zwillinger1999crc} and the Pearson Linear Correlation Coefficient (PLCC)~\cite{dekking2005modern} were employed in our experiments. SRCC reflects the strength and direction of the monotonic relationship between predicted and actual scores, while PLCC captures their linear correlation; both range from -1 to 1, with higher values denoting stronger alignment.
To achieve more comprehensive benchmarking, we also include Kendall’s Rank Correlation Coefficient (KRCC)~\cite{kendall1938new}, which quantifies the proportion of correctly ordered pairs. This provides a more robust measure of pairwise agreement, especially when dealing with ties or small sample sizes. Together, these three metrics offer a thorough assessment of model accuracy and its alignment with human perceptual judgments. Given that comparisons of perceptual scores yield relevance only when conducted within individual scenes~\cite{liang2024perceptual}, the results aggregated across the dataset are presented as the mean and standard deviation of the results calculated on a scene-by-scene basis. Additionally, detailed scenewise results are delineated in the subsequent sections for comprehensive analysis.

\noindent{\textbf{Training setup.}}
The model was trained using the ADAM optimizer~\cite{kingma2014adam} for 200 epochs with a batch size of 16, resulting in a total training time of approximately 56.99 hours. Once training on the unlabeled dataset is complete, the model can be efficiently applied to novel scenes across varied datasets using only minor linear regression. The computational experiments were conducted on a high-specification desktop equipped with an AMD 5950X processor, an RTX 4090 GPU, and 128 GB of RAM, running on Windows 10 operating system. The implementation was carried out in PyTorch~\cite{paszke2019pytorch}.

\begin{table*}[!htb]
\centering
\caption{\textbf{Quantitative evaluation of various no-reference quality assessment methods across the Fieldwork, LLFF, and Lab datasets}, including means and standard deviations of SRCC, PLCC, and KRCC. For each column, the best results are highlighted in bold, with the concluding row indicating the enhancement relative to the second-best result.}
\label{tab:benchmarking}
\scriptsize
\renewcommand{\arraystretch}{1.2}
\setlength{\tabcolsep}{2pt}
\begin{tabular}{c|rrr|rrr|rrr}
\hline
\hline
&\multicolumn{3}{c}{Fieldwork}&\multicolumn{3}{c}{LLFF}&\multicolumn{3}{c}{Lab}\\
\hline
{Method} & {SRCC ↑ (std)} & {PLCC ↑ (std)} & {KRCC ↑ (std)} & {SRCC ↑ (std)} & {PLCC ↑ (std)} & {KRCC ↑ (std)} & {SRCC ↑ (std)} & {PLCC ↑ (std)} & {KRCC ↑ (std)} \\
\hline

TV & \phantom{+}$0.378$ ($0.64$) & \phantom{+}$0.423$ ($0.58$) & \phantom{+}$0.311$ ($0.62$) & \phantom{+}$0.087$ ($0.65$) & \phantom{+}$0.050$ ($0.61$) & \phantom{+}$0.075$ ($0.52$) & \phantom{+}$0.200$ ($0.48$) & \phantom{+}$0.136$ ($0.30$) & \phantom{+}$\underline{0.229}$ ($0.36$)\\
BRISQUE & \phantom{+}$0.089$ ($0.58$) & \phantom{+}$0.152$ ($0.58$) & \phantom{+}$0.067$ ($0.53$) & $-0.037$ ($0.54$) & $-0.103$ ($0.45$) & $-0.050$ ($0.44$) & \phantom{+}$\underline{0.214}$ ($0.83$) & \phantom{+}$0.204$ ($0.72$) & \phantom{+}$0.171$ ($0.70$)\\
NIQE & \phantom{+}$0.467$ ($0.63$) & \phantom{+}$0.331$ ($0.66$) & \phantom{+}$0.400$ ($0.56$) & \phantom{+}$0.025$ ($0.50$) & $-0.077$ ($0.43$) & \phantom{+}$0.025$ ($0.42$) & $-0.357$ ($0.16$) & $-0.329$ ($0.12$) & $-0.314$ ($0.17$)\\
PIQE & \phantom{+}$0.079$ ($0.49$) & $-0.079$ ($0.55$) & \phantom{+}$0.273$ ($0.57$) & \phantom{+}$0.047$ ($0.56$) & $-0.079$ ($0.47$) & $-0.021$ ($0.44$) & $-0.364$ ($0.65$) & $-0.060$ ($0.70$) & $-0.302$ ($0.49$)\\
CLIP-IQA & \phantom{+}$0.233$ ($0.63$) & \phantom{+}$0.178$ ($0.57$) & \phantom{+}$0.200$ ($0.54$) & \phantom{+}$0.025$ ($0.47$) & $-0.046$ ($0.36$) & \phantom{+}$0.025$ ($0.34$) & $-0.057$ ($0.50$) & $-0.240$ ($0.47$) & $-0.086$ ($0.36$)\\
CONTRIQUE & \phantom{+}$\underline{0.689}$ ($0.28$) & \phantom{+}$\underline{0.759}$ ($0.29$) & \phantom{+}$\underline{0.622}$ ($0.29$) & \phantom{+}$0.350$ ($0.40$) & \phantom{+}$0.400$ ($0.50$) & \phantom{+}$0.275$ ($0.33$) & \phantom{+}$0.086$ ($0.43$) & \phantom{+}$0.200$ ($0.53$) & \phantom{+}$0.057$ ($0.36$)\\
Re-IQA & \phantom{+}$0.589$ ($0.53$) & \phantom{+}$0.585$ ($0.48$) & \phantom{+}$0.489$ ($0.44$) & \phantom{+}$0.062$ ($0.71$) & $-0.018$ ($0.71$) & \phantom{+}$0.025$ ($0.61$) & \phantom{+}$0.143$ ($0.16$) & \phantom{+}$\underline{0.213}$ ($0.30$) & \phantom{+}$0.200$ ($0.10$)\\
\hline
VIIDEO & \phantom{+}$0.022$ ($0.37$) & \phantom{+}$0.070$ ($0.44$) & \phantom{+}$0.022$ ($0.28$) & $-0.050$ ($0.49$) & \phantom{+}$0.002$ ($0.49$) & $-0.000$ ($0.39$) & \phantom{+}$0.000$ ($0.27$) & \phantom{+}$0.061$ ($0.47$) & $-0.029$ ($0.30$)\\
Video-BlIINDS & \phantom{+}$0.189$ ($0.39$) & \phantom{+}$0.148$ ($0.41$) & \phantom{+}$0.156$ ($0.31$) & \phantom{+}$0.162$ ($0.44$) & \phantom{+}$0.051$ ($0.40$) & \phantom{+}$0.175$ ($0.38$) & $-0.314$ ($0.33$) & $-0.113$ ($0.42$) & $-0.286$ ($0.33$)\\
FAST-VQA & \phantom{+}$0.167$ ($0.59$) & \phantom{+}$0.106$ ($0.59$) & \phantom{+}$0.122$ ($0.46$) & \phantom{+}$0.112$ ($0.55$) & \phantom{+}$0.255$ ($0.55$) & \phantom{+}$0.185$ ($0.41$) & $-0.171$ ($0.65$) & $-0.191$ ($0.63$) & $-0.186$ ($0.52$)\\
FasterVQA & \phantom{+}$0.186$ ($0.51$) & \phantom{+}$0.245$ ($0.55$) & \phantom{+}$0.172$ ($0.47$) & \phantom{+}$0.162$ ($0.58$) & \phantom{+}$0.176$ ($0.49$) & \phantom{+}$0.125$ ($0.48$) & $-0.129$ ($0.46$) & $-0.175$ ($0.44$) & $-0.186$ ($0.38$)\\
DOVER & \phantom{+}$0.200$ ($0.56$) & \phantom{+}$0.267$ ($0.65$) & \phantom{+}$0.267$ ($0.50$) & \phantom{+}$0.150$ ($0.61$) & \phantom{+}$0.153$ ($0.47$) & \phantom{+}$0.125$ ($0.48$) & $-0.129$ ($0.26$) & $-0.209$ ($0.18$) & $-0.214$ ($0.22$)\\
DOVER-Mobile & \phantom{+}$0.344$ ($0.60$) & \phantom{+}$0.341$ ($0.61$) & \phantom{+}$0.311$ ($0.47$) & \phantom{+}$0.200$ ($0.47$) & \phantom{+}$0.254$ ($0.36$) & \phantom{+}$0.125$ ($0.40$) & $-0.071$ ($0.47$) & $-0.097$ ($0.32$) & $-0.143$ ($0.46$)\\
\hline
NR-LFQA & \phantom{+}$0.207$ ($0.52$) & \phantom{+}$0.157$ ($0.56$) & \phantom{+}$0.149$ ($0.42$) & \phantom{+}$0.088$ ($0.47$) & \phantom{+}$0.106$ ($0.34$) & \phantom{+}$0.114$ ($0.37$) & $-0.094$ ($0.59$) & $-0.067$ ($0.57$) & $-0.039$ ($0.51$)\\
Tensor-NLFQ & \phantom{+}$0.282$ ($0.42$) & \phantom{+}$0.290$ ($0.49$) & \phantom{+}$0.202$ ($0.43$) & \phantom{+}$0.223$ ($0.38$) & \phantom{+}$0.210$ ($0.36$) & \phantom{+}$0.125$ ($0.29$) & $-0.061$ ($0.64$) & $-0.064$ ($0.57$) & $-0.072$ ($0.55$)\\
ALAS-DADS & \phantom{+}$0.356$ ($0.49$) & \phantom{+}$0.183$ ($0.54$) & \phantom{+}$0.244$ ($0.40$) & \phantom{+}$0.138$ ($0.35$) & \phantom{+}$0.125$ ($0.34$) & \phantom{+}$0.050$ ($0.38$) & \phantom{+}$0.043$ ($0.59$) & \phantom{+}$0.166$ ($0.59$) & \phantom{+}$0.029$ ($0.48$)\\
LFACon & \phantom{+}$0.333$ ($0.44$) & \phantom{+}$0.310$ ($0.47$) & \phantom{+}$0.242$ ($0.33$) & \phantom{+}$\underline{0.412}$ ($0.48$) & \phantom{+}$\underline{0.395}$ ($0.31$) & \phantom{+}$\underline{0.325}$ ($0.36$) & \phantom{+}$0.157$ ($0.60$) & \phantom{+}$0.132$ ($0.53$) & \phantom{+}$0.086$ ($0.51$)\\

\hline
\textbf{Proposed}  & \phantom{+}$\pmb{ 0.911 }$  ($0.08$) & \phantom{+}$\pmb{ 0.883 }$  ($0.11$) & \phantom{+}$\pmb{ 0.822 }$  ($0.16$) & \phantom{+}$\pmb{ 0.700 }$  ($0.28$) & \phantom{+}$\pmb{ 0.644 }$  ($0.39$) & \phantom{+}$\pmb{ 0.625 }$  ($0.19$) & \phantom{+}$\pmb{ 0.700 }$  ($0.04$) & \phantom{+}$\pmb{ 0.678 }$  ($0.10$) & \phantom{+}$\pmb{ 0.571 }$  ($0.09$)\\
\hline
\multirow{2}{*}{ \textbf{V.S. 2nd Best} }  & \multicolumn{1}{c}{ $\pmb{ +0.222 }$ } & \multicolumn{1}{c}{ $ \pmb{ +0.124 } $ } & \multicolumn{1}{c|}{ $ \pmb{ +0.200 } $ } & \multicolumn{1}{c}{ $ \pmb{ +0.287 } $ } & \multicolumn{1}{c}{ $ \pmb{ +0.245 } $ } & \multicolumn{1}{c|}{ $ \pmb{ +0.300 } $ } & \multicolumn{1}{c}{ $ \pmb{ +0.486 } $ } & \multicolumn{1}{c}{ $ \pmb{ +0.465 } $ } & \multicolumn{1}{c}{ $ \pmb{ +0.343 } $ }\\ & \multicolumn{1}{c}{ $ \pmb{ +32.3\% } $ } & \multicolumn{1}{c}{ $ \pmb{ +16.4\% } $ } & \multicolumn{1}{c|}{ $ \pmb{ +32.1\% } $ } & \multicolumn{1}{c}{ $ \pmb{ +69.7\% } $ } & \multicolumn{1}{c}{ $ \pmb{ +61.2\% } $ } & \multicolumn{1}{c|}{ $ \pmb{ +92.3\% } $ } & \multicolumn{1}{c}{ $ \pmb{ +226.7\% } $ } & \multicolumn{1}{c}{ $ \pmb{ +218.3\% } $ } & \multicolumn{1}{c}{ $ \pmb{ +150.0\% } $ }\\

\hline
\hline
\end{tabular}
\end{table*}

\begin{figure*}[htb]
\centering
\includegraphics[width=\textwidth]{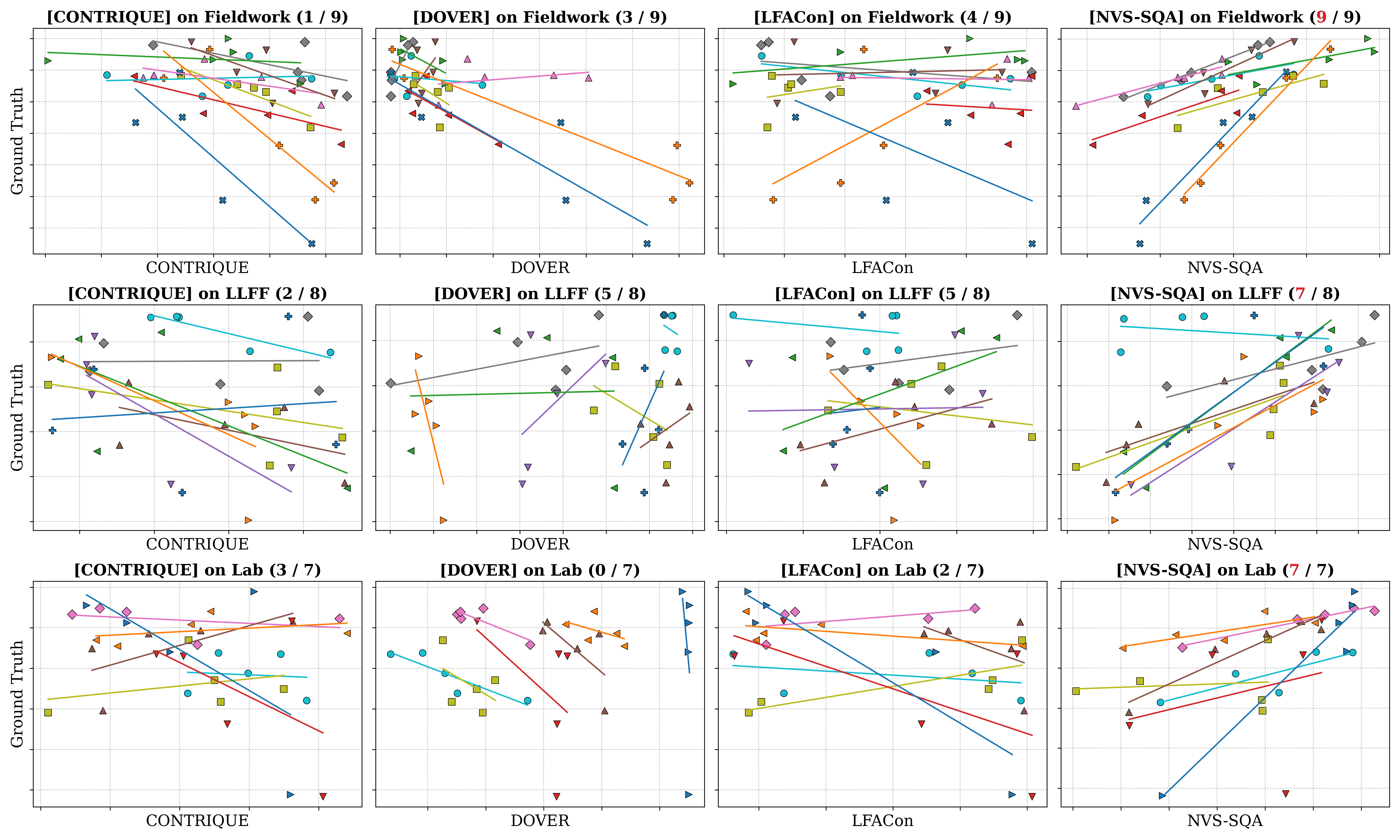}
\caption{
\textbf{How often does each quality estimation method align with ground-truth perceptual labels (indicated by /'-shaped lines)?} We compare our proposed NVS-SQA against three competitive representatives of no-reference quality assessment: CONTRIQUE (IQA), DOVER (VQA), and LFACon (LFIQA), evaluated on the Fieldwork, LLFF, and Lab datasets. Each subfigure shows scatter plots where different symbols and colors represent individual scenes, with a linear regression line for each. Each subfigure reports the number of scenes within the dataset where the predicted scores exhibit a positive correlation (positive slope) with human perceptual scores. For example, '9/9' in Fieldwork indicates that NVS-SQA achieved positive alignment in all 9 individual scenes.}

\label{fig:bench_plots}
\end{figure*}

\subsection{Comparison with no-reference quality assessment methods}

We benchmarked our approach against a diverse set of no-reference quality assessment methods spanning multiple domains, including IQA (TV~\cite{bahrami2016efficient}, BRISQUE~\cite{mittal2012no}, NIQE~\cite{mittal2012making}, PIQE~\cite{venkatanath2015blind}, CLIP-IQA~\cite{wang2023exploring}, CONTRIQUE~\cite{madhusudana2022image}, Re-IQA~\cite{saha2023re}), VQA (VIIDEO~\cite{mittal2015completely}, Video-BLIINDS~\cite{saad2014blind}, FAST-VQA~\cite{wu2022fast}, FasterVQA~\cite{wu2023neighbourhood}, DOVER~\cite{wu2023dover}, DOVER-Mobile~\cite{wu2023dover}), and LFIQA (NR-LFQA~\cite{shi2019no}, Tensor-NLFQ~\cite{zhou2020tensor}, ALAS-DADS~\cite{qu2021light}, LFACon~\cite{qu2023lfacon}).
Table~\ref{tab:benchmarking} presents the evaluation results (SRCC, PLCC, and KRCC) for these no-reference quality assessment methods, including our NVS-SQA framework, across the Fieldwork, LLFF, and Lab datasets. Optimal performances within each dataset are emphasized in bold, and the table's final row delineates the NVS-SQA's relative performance enhancement over the second-ranking method. The analysis underscores the superior performance of NVS-SQA across all datasets. Within the Fieldwork dataset, NVS-SQA achieved an increment of 32.3\% in the SRCC, amounting to 0.222, relative to its nearest competitor. In the LLFF dataset, it demonstrated an improvement, with a 69.7\% increase in SRCC (0.287) and a 92.3\% rise in the KRCC (0.300), compared to the second-best method. Similarly, for the Lab dataset, NVS-SQA recorded an enhancement, with SRCC increasing by 226\% (0.486) and PLCC by 218\% (0.465). Moreover, NVS-SQA exhibits the smallest standard deviation across most of the metrics, demonstrating its stability relative to other methods.

\noindent{\textbf{Visualiazation of correlation with perceptual labels.}}
Figure~\ref{fig:bench_plots} illustrates how well the predicted quality assessments correlate with ground-truth perceptual labels. The ground-truth perceptual labels used in this analysis are the JOD scores described in Section~\ref{subsec:multibranch_contrastive_objective}. These labels reflect human-perceived quality differences and were computed from pairwise comparison data as part of the public datasets in~\cite{liang2024perceptual}. Each subplot corresponds to a dataset, and each regression line represents one individual scene within that dataset. The count (e.g., 9/9 for Fieldwork) reflects the number of scenes for which the model's predicted quality scores positively correlate with human labels. This provides a scene-level consistency measure beyond global correlation coefficients. As shown in the figure, NVS-SQA achieves the highest number of positively sloped lines in every dataset, reaching perfect alignment in Fieldwork (9 of 9 lines) and top performance in both LLFF (7 of 8) and Lab (7 of 7). This result underscores the robust effectiveness of NVS-SQA across a diverse set of scenes and highlights its superior alignment with human perception compared to existing approaches.

\noindent{\textbf{Cross-dataset evaluation.}}
To further assess the generalization of the proposed method, we conducted a cross-dataset evaluation, the results of which are detailed in Table~\ref{tab:cross_dataset}. Specifically, in this setup, the model is regressed on two datasets (A and B) and then tested on the remaining one (C). For example, the Fieldwork results are obtained by regressing on LLFF and Lab. Comparing these results with those from Table~\ref{tab:benchmarking} indicates that the performance of the proposed method does not plunge during cross-validation, and even further widens the performance gap with the second-best in some cases. In two datasets, the method's performance relative to the second-best quality assessment method shows even greater improvement. Specifically, in the Fieldwork dataset, NVS-SQA achieved an increase of 39.4\% in the SRCC and a 44.8\% enhancement in the KRCC compared to the second best. Similarly, in the LLFF dataset, it recorded a 215\% rise in KLCC.

\begin{table*}[!htb]
\centering
\caption{\textbf{Cross-dataset evaluation against various no-reference quality assessment methods}, using measures including means and standard deviations of SRCC, PLCC, and KRCC. In this cross-dataset setup, the model is regressed on two datasets (A and B) and then tested on the remaining one (C); for instance, the Fieldwork results derive from training on LLFF and Lab. For each column, the best results are highlighted in bold.}
\label{tab:cross_dataset}
\scriptsize
\renewcommand{\arraystretch}{1.2}
\setlength{\tabcolsep}{2pt}
\begin{tabular}{c|rrr|rrr|rrr}
\hline
\hline
&\multicolumn{3}{c}{Fieldwork}&\multicolumn{3}{c}{LLFF}&\multicolumn{3}{c}{Lab}\\
\hline
{Method} & {SRCC ↑ (std)} & {PLCC ↑ (std)} & {KRCC ↑ (std)} & {SRCC ↑ (std)} & {PLCC ↑ (std)} & {KRCC ↑ (std)} & {SRCC ↑ (std)} & {PLCC ↑ (std)} & {KRCC ↑ (std)} \\
\hline

TV & \phantom{+}$0.344$ ($0.58$) & \phantom{+}$0.398$ ($0.58$) & \phantom{+}$0.244$ ($0.54$) & $-0.050$ ($0.50$) & $-0.032$ ($0.53$) & \phantom{+}$0.025$ ($0.42$) & \phantom{+}$0.143$ ($0.51$) & \phantom{+}$0.266$ ($0.27$) & \phantom{+}$0.171$ ($0.46$)\\
BRISQUE & $-0.044$ ($0.59$) & $-0.004$ ($0.58$) & $-0.067$ ($0.51$) & $-0.237$ ($0.50$) & $-0.289$ ($0.52$) & $-0.175$ ($0.42$) & \phantom{+}$\underline{0.214}$ ($0.42$) & \phantom{+}$0.279$ ($0.43$) & \phantom{+}$0.171$ ($0.35$)\\
NIQE & \phantom{+}$0.344$ ($0.37$) & \phantom{+}$0.337$ ($0.40$) & \phantom{+}$0.311$ ($0.33$) & $-0.212$ ($0.58$) & $-0.138$ ($0.55$) & $-0.175$ ($0.50$) & $-0.243$ ($0.22$) & $-0.188$ ($0.41$) & $-0.171$ ($0.22$)\\
PIQE & $-0.092$ ($0.51$) & $-0.012$ ($0.58$) & $-0.211$ ($0.56$) & $-0.175$ ($0.55$) & $-0.173$ ($0.51$) & $-0.171$ ($0.54$) & $-0.169$ ($0.30$) & $-0.237$ ($0.25$) & $-0.295$ ($0.21$)\\
CLIP-IQA & \phantom{+}$0.444$ ($0.34$) & \phantom{+}$0.525$ ($0.32$) & \phantom{+}$0.356$ ($0.25$) & \phantom{+}$0.175$ ($0.60$) & \phantom{+}$0.264$ ($0.67$) & \phantom{+}$0.100$ ($0.49$) & $-0.057$ ($0.25$) & \phantom{+}$0.020$ ($0.26$) & $-0.029$ ($0.22$)\\
CONTRIQUE & \phantom{+}$\underline{0.622}$ ($0.38$) & \phantom{+}$\underline{0.626}$ ($0.31$) & \phantom{+}$\underline{0.522}$ ($0.42$) & $-0.013$ ($0.34$) & \phantom{+}$0.050$ ($0.48$) & $-0.075$ ($0.20$) & $-0.143$ ($0.33$) & $-0.102$ ($0.56$) & $-0.114$ ($0.32$)\\
Re-IQA & \phantom{+}$0.578$ ($0.51$) & \phantom{+}$0.614$ ($0.40$) & \phantom{+}$0.511$ ($0.50$) & \phantom{+}$0.050$ ($0.51$) & \phantom{+}$0.052$ ($0.54$) & \phantom{+}$0.050$ ($0.46$) & \phantom{+}$0.043$ ($0.40$) & \phantom{+}$0.116$ ($0.48$) & \phantom{+}$0.000$ ($0.30$)\\
\hline
VIIDEO & \phantom{+}$0.222$ ($0.54$) & \phantom{+}$0.242$ ($0.45$) & \phantom{+}$0.178$ ($0.42$) & $-0.412$ ($0.46$) & $-0.437$ ($0.35$) & $-0.325$ ($0.36$) & \phantom{+}$0.143$ ($0.52$) & $-0.034$ ($0.56$) & \phantom{+}$0.086$ ($0.44$)\\
Video-BlIINDS & \phantom{+}$0.067$ ($0.40$) & \phantom{+}$0.107$ ($0.41$) & \phantom{+}$0.067$ ($0.34$) & \phantom{+}$0.088$ ($0.59$) & \phantom{+}$0.206$ ($0.57$) & \phantom{+}$0.100$ ($0.50$) & \phantom{+}$0.186$ ($0.53$) & \phantom{+}$\underline{0.299}$ ($0.50$) & \phantom{+}$0.171$ ($0.43$)\\
FAST-VQA & \phantom{+}$0.063$ ($0.56$) & \phantom{+}$0.030$ ($0.57$) & \phantom{+}$0.022$ ($0.43$) & \phantom{+}$0.062$ ($0.65$) & $-0.002$ ($0.65$) & \phantom{+}$0.075$ ($0.53$) & $-0.286$ ($0.45$) & $-0.288$ ($0.38$) & $-0.257$ ($0.45$)\\
FasterVQA & \phantom{+}$0.104$ ($0.54$) & \phantom{+}$0.129$ ($0.62$) & \phantom{+}$0.100$ ($0.45$) & \phantom{+}$\underline{0.412}$ ($0.46$) & \phantom{+}$\underline{0.486}$ ($0.36$) & \phantom{+}$\underline{0.375}$ ($0.42$) & \phantom{+}$-0.286$ ($0.32$) & $-0.160$ ($0.38$) & $-0.114$ ($0.30$)\\
DOVER & \phantom{+}$0.189$ ($0.68$) & \phantom{+}$0.190$ ($0.66$) & \phantom{+}$0.144$ ($0.62$) & \phantom{+}$0.112$ ($0.57$) & \phantom{+}$0.290$ ($0.56$) & \phantom{+}$0.075$ ($0.46$) & $-0.129$ ($0.27$) & $-0.143$ ($0.37$) & $-0.186$ ($0.22$)\\
DOVER-Mobile & \phantom{+}$0.367$ ($0.42$) & \phantom{+}$0.449$ ($0.40$) & \phantom{+}$0.311$ ($0.39$) & \phantom{+}$0.262$ ($0.49$) & \phantom{+}$0.268$ ($0.51$) & \phantom{+}$0.250$ ($0.40$) & $-0.229$ ($0.27$) & $-0.175$ ($0.25$) & $-0.186$ ($0.24$)\\
\hline
NR-LFQA & \phantom{+}$0.168$ ($0.50$) & \phantom{+}$0.130$ ($0.53$) & \phantom{+}$0.120$ ($0.46$) & \phantom{+}$0.052$ ($0.44$) & \phantom{+}$0.066$ ($0.42$) & \phantom{+}$0.066$ ($0.38$) & $-0.102$ ($0.57$) & $-0.093$ ($0.52$) & $-0.061$ ($0.55$)\\
Tensor-NLFQ & \phantom{+}$0.178$ ($0.48$) & \phantom{+}$0.201$ ($0.42$) & \phantom{+}$0.159$ ($0.40$) & \phantom{+}$0.164$ ($0.35$) & \phantom{+}$0.139$ ($0.38$) & \phantom{+}$0.071$ ($0.33$) & $-0.048$ ($0.58$) & $-0.027$ ($0.54$) & $-0.081$ ($0.52$)\\
ALAS-DADS & \phantom{+}$0.144$ ($0.46$) & \phantom{+}$0.146$ ($0.53$) & \phantom{+}$0.156$ ($0.37$) & \phantom{+}$0.175$ ($0.54$) & \phantom{+}$0.226$ ($0.52$) & \phantom{+}$0.125$ ($0.40$) & $-0.029$ ($0.19$) & \phantom{+}$0.054$ ($0.25$) & $-0.029$ ($0.17$)\\
LFACon & \phantom{+}$0.233$ ($0.45$) & \phantom{+}$0.182$ ($0.45$) & \phantom{+}$0.122$ ($0.42$) & \phantom{+}$0.350$ ($0.36$) & \phantom{+}$0.385$ ($0.46$) & \phantom{+}$0.200$ ($0.35$) & \phantom{+}$0.143$ ($0.53$) & \phantom{+}$0.212$ ($0.50$) & \phantom{+}$\underline{0.186}$ ($0.45$)\\

\hline
\textbf{Proposed}  & \phantom{+}$\pmb{ 0.867 }$  ($0.08$) & \phantom{+}$\pmb{ 0.812 }$  ($0.24$) & \phantom{+}$\pmb{ 0.756 }$  ($0.12$) & \phantom{+}$\pmb{ 0.671 }$  ($0.12$) & \phantom{+}$\pmb{ 0.647 }$  ($0.23$) & \phantom{+}$\pmb{ 0.565 }$  ($0.25$) & \phantom{+}$\pmb{ 0.670 }$  ($0.09$) & \phantom{+}$\pmb{ 0.669 }$  ($0.18$) & \phantom{+}$\pmb{ 0.586 }$  ($0.26$)\\
\hline\multirow{2}{*}{ \textbf{V.S. 2nd Best} }  & \multicolumn{1}{c}{ $ \pmb{ +0.245 } $ } & \multicolumn{1}{c}{ $ \pmb{ +0.186 } $ } & \multicolumn{1}{c}{ $ \pmb{ +0.234 } $ } & \multicolumn{1}{c}{ $ \pmb{ +0.259 } $ } & \multicolumn{1}{c}{ $ \pmb{ +0.161 } $ } & \multicolumn{1}{c}{ $ \pmb{ +0.190 } $ } & \multicolumn{1}{c}{ $ \pmb{ +0.456 } $ } & \multicolumn{1}{c}{ $ \pmb{ +0.370 } $ } & \multicolumn{1}{c}{ $ \pmb{ +0.400 } $ }\\ & \multicolumn{1}{c}{ $ \pmb{ +39.4\% } $ } & \multicolumn{1}{c}{ $ \pmb{ +29.7\% } $ } & \multicolumn{1}{c}{ $ \pmb{ +44.8\% } $ } & \multicolumn{1}{c}{ $ \pmb{ +62.9\% } $ } & \multicolumn{1}{c}{ $ \pmb{ +33.1\% } $ } & \multicolumn{1}{c}{ $ \pmb{ +50.7\% } $ } & \multicolumn{1}{c}{ $ \pmb{ +213.1\% } $ } & \multicolumn{1}{c}{ $ \pmb{ +123.7\% } $ } & \multicolumn{1}{c}{ $ \pmb{ +215.1\% } $ }\\

\hline
\hline
\end{tabular}
\end{table*}

\begin{table*}[htb]
\centering
\caption{\textbf{Comparison of the proposed no-reference method against prevalent full-reference quality assessment methods}, including means and standard deviations of SRCC, PLCC, and KRCC. The highest scores in each column are bold. \textbf{Due to the absence of reference videos in the LLFF dataset, the results for STRRED, VMAF, and FovVideoVDP are denoted as "–".}}
\label{tab:fr_benchmarking}
\scriptsize
\renewcommand{\arraystretch}{1.2}
\setlength{\tabcolsep}{2pt}
\begin{tabular}{c|rrr|rrr|rrr}
\hline
\hline
&\multicolumn{3}{c}{Fieldwork}&\multicolumn{3}{c}{LLFF}&\multicolumn{3}{c}{Lab}\\
\hline
{Method} & {SRCC ↑ (std)} & {PLCC ↑ (std)} & {KRCC ↑ (std)} & {SRCC ↑ (std)} & {PLCC ↑ (std)} & {KRCC ↑ (std)} & {SRCC ↑ (std)} & {PLCC ↑ (std)} & {KRCC ↑ (std)} \\
\hline

PSNR & \phantom{+}$0.611$ ($0.36$) & \phantom{+}$0.629$ ($0.36$) & \phantom{+}$0.489$ ($0.38$) & \phantom{+}$0.050$ ($0.56$) & \phantom{+}$0.051$ ($0.58$) & \phantom{+}$0.050$ ($0.49$) & \phantom{+}$0.200$ ($0.71$) & \phantom{+}$0.160$ ($0.73$) & \phantom{+}$0.143$ ($0.65$)\\
SSIM & \phantom{+}$0.733$ ($0.26$) & \phantom{+}$0.729$ ($0.22$) & \phantom{+}$0.622$ ($0.28$) & \phantom{+}$0.400$ ($0.49$) & \phantom{+}$0.488$ ($0.50$) & \phantom{+}$0.350$ ($0.37$) & \phantom{+}$0.529$ ($0.23$) & \phantom{+}$0.358$ ($0.36$) & \phantom{+}$0.429$ ($0.22$)\\
MS-SSIM & \phantom{+}$0.800$ ($0.17$) & \phantom{+}$0.742$ ($0.22$) & \phantom{+}$0.689$ ($0.22$) & \phantom{+}$0.375$ ($0.48$) & \phantom{+}$0.447$ ($0.54$) & \phantom{+}$0.325$ ($0.36$) & \phantom{+}$0.571$ ($0.11$) & \phantom{+}$0.426$ ($0.36$) & \phantom{+}$0.457$ ($0.14$)\\
IW-SSIM & \phantom{+}$0.800$ ($0.12$) & \phantom{+}$0.721$ ($0.22$) & \phantom{+}$0.644$ ($0.16$) & \phantom{+}$0.375$ ($0.48$) & \phantom{+}$0.424$ ($0.53$) & \phantom{+}$0.325$ ($0.36$) & \phantom{+}$0.600$ ($0.04$) & \phantom{+}$0.491$ ($0.37$) & \phantom{+}$0.514$ ($0.09$)\\
VIF & \phantom{+}$0.778$ ($0.28$) & \phantom{+}$0.704$ ($0.21$) & \phantom{+}$0.711$ ($0.26$) & \phantom{+}$0.225$ ($0.53$) & \phantom{+}$0.265$ ($0.57$) & \phantom{+}$0.200$ ($0.40$) & \phantom{+}$0.214$ ($0.69$) & \phantom{+}$0.131$ ($0.56$) & \phantom{+}$0.171$ ($0.57$)\\
FSIM & \phantom{+}$\underline{0.822}$ ($0.19$) & \phantom{+}$\underline{0.785}$ ($0.23$) & \phantom{+}$0.711$ ($0.20$) & \phantom{+}$0.387$ ($0.49$) & \phantom{+}$0.458$ ($0.52$) & \phantom{+}$0.350$ ($0.37$) & \phantom{+}$0.500$ ($0.41$) & \phantom{+}$0.416$ ($0.37$) & \phantom{+}$0.429$ ($0.47$)\\
GMSD & \phantom{+}$0.778$ ($0.33$) & \phantom{+}$0.740$ ($0.31$) & \phantom{+}$0.711$ ($0.28$) & \phantom{+}$0.387$ ($0.48$) & \phantom{+}$0.398$ ($0.55$) & \phantom{+}$0.350$ ($0.40$) & \phantom{+}$0.514$ ($0.16$) & \phantom{+}$0.408$ ($0.35$) & \phantom{+}$0.457$ ($0.22$)\\
VSI & \phantom{+}$0.767$ ($0.37$) & \phantom{+}$0.699$ ($0.43$) & \phantom{+}$0.689$ ($0.36$) & \phantom{+}$0.350$ ($0.53$) & \phantom{+}$0.412$ ($0.55$) & \phantom{+}$0.300$ ($0.41$) & \phantom{+}$0.614$ ($0.07$) & \phantom{+}$0.482$ ($0.28$) & \phantom{+}$0.543$ ($0.14$)\\
DSS & \phantom{+}$0.789$ ($0.33$) & \phantom{+}$0.766$ ($0.27$) & \phantom{+}$0.711$ ($0.28$) & \phantom{+}$\underline{0.462}$ ($0.48$) & \phantom{+}$\underline{0.493}$ ($0.48$) & \phantom{+}$\underline{0.425}$ ($0.41$) & \phantom{+}$0.257$ ($0.72$) & \phantom{+}$0.221$ ($0.74$) & \phantom{+}$0.171$ ($0.64$)\\
HaarPSI & \phantom{+}$0.811$ ($0.34$) & \phantom{+}$0.747$ ($0.28$) & \phantom{+}$\underline{0.756}$ ($0.28$) & \phantom{+}$0.362$ ($0.47$) & \phantom{+}$0.425$ ($0.52$) & \phantom{+}$0.300$ ($0.36$) & \phantom{+}$0.343$ ($0.36$) & \phantom{+}$0.365$ ($0.32$) & \phantom{+}$0.286$ ($0.38$)\\
MDSI & \phantom{+}$0.722$ ($0.36$) & \phantom{+}$0.713$ ($0.31$) & \phantom{+}$0.644$ ($0.32$) & \phantom{+}$0.437$ ($0.49$) & \phantom{+}$0.424$ ($0.54$) & \phantom{+}$0.400$ ($0.41$) & \phantom{+}$0.200$ ($0.72$) & \phantom{+}$0.144$ ($0.70$) & \phantom{+}$0.143$ ($0.62$)\\
LPIPS & \phantom{+}$0.722$ ($0.26$) & \phantom{+}$0.628$ ($0.27$) & \phantom{+}$0.622$ ($0.26$) & \phantom{+}$0.187$ ($0.58$) & \phantom{+}$0.262$ ($0.62$) & \phantom{+}$0.175$ ($0.46$) & $-0.300$ ($0.61$) & $-0.129$ ($0.65$) & $-0.286$ ($0.54$)\\
DISTS & \phantom{+}$0.789$ ($0.17$) & \phantom{+}$0.720$ ($0.17$) & \phantom{+}$0.667$ ($0.17$) & \phantom{+}$0.337$ ($0.50$) & \phantom{+}$0.428$ ($0.49$) & \phantom{+}$0.300$ ($0.37$) & \phantom{+}$0.557$ ($0.28$) & \phantom{+}$\underline{0.593}$ ($0.07$) & \phantom{+}$0.486$ ($0.30$)\\
STRRED & \phantom{+}$0.800$ ($0.24$) & \phantom{+}$0.767$ ($0.30$) & \phantom{+}$0.711$ ($0.26$) & $-$ & $-$ & $-$ & \phantom{+}$0.543$ ($0.14$) & \phantom{+}$0.491$ ($0.31$) & \phantom{+}$0.429$ ($0.22$)\\
VMAF & \phantom{+}$0.556$ ($0.50$) & \phantom{+}$0.551$ ($0.49$) & \phantom{+}$0.511$ ($0.46$) & $-$ & $-$ & $-$ & \phantom{+}$0.414$ ($0.13$) & \phantom{+}$0.474$ ($0.16$) & \phantom{+}$0.286$ ($0.17$)\\
FovVideoVDP & \phantom{+}$0.789$ ($0.33$) & \phantom{+}$0.780$ ($0.22$) & \phantom{+}$0.733$ ($0.36$) & $-$ & $-$ & $-$ & \phantom{+}$\underline{0.657}$ ($0.07$) & \phantom{+}$0.472$ ($0.30$) & \phantom{+}$\pmb{0.571}$ ($0.14$)\\

\hline
\textbf{Proposed}  & \phantom{+}$\pmb{ 0.911 }$  ($0.08$) & \phantom{+}$\pmb{ 0.883 }$  ($0.11$) & \phantom{+}$\pmb{ 0.822 }$  ($0.16$) & \phantom{+}$\pmb{ 0.700 }$  ($0.28$) & \phantom{+}$\pmb{ 0.644 }$  ($0.39$) & \phantom{+}$\pmb{ 0.625 }$  ($0.19$) & \phantom{+}$\pmb{ 0.700 }$  ($0.04$) & \phantom{+}$\pmb{ 0.678 }$  ($0.10$) & \phantom{+}$\pmb{ 0.571 }$  ($0.09$)\\
\hline\multirow{2}{*}{ \textbf{V.S. 2nd Best} }  & \multicolumn{1}{c}{ $ \pmb{ +0.089 } $ } & \multicolumn{1}{c}{ $ \pmb{ +0.098 } $ } & \multicolumn{1}{c}{ $ \pmb{ +0.067 } $ } & \multicolumn{1}{c}{ $ \pmb{ +0.237 } $ } & \multicolumn{1}{c}{ $ \pmb{ +0.151 } $ } & \multicolumn{1}{c}{ $ \pmb{ +0.200 } $ } & \multicolumn{1}{c}{ $ \pmb{ +0.043 } $ } & \multicolumn{1}{c}{ $ \pmb{ +0.085 } $ } & \multicolumn{1}{c}{ $ \pmb{ +0.000 } $ }\\ & \multicolumn{1}{c}{ $ \pmb{ +10.8\% } $ } & \multicolumn{1}{c}{ $ \pmb{ +12.5\% } $ } & \multicolumn{1}{c}{ $ \pmb{ +8.8\% } $ } & \multicolumn{1}{c}{ $ \pmb{ +51.4\% } $ } & \multicolumn{1}{c}{ $ \pmb{ +30.6\% } $ } & \multicolumn{1}{c}{ $ \pmb{ +47.1\% } $ } & \multicolumn{1}{c}{ $ \pmb{ +6.5\% } $ } & \multicolumn{1}{c}{ $ \pmb{ +14.4\% } $ } & \multicolumn{1}{c}{ $ \pmb{ +0.0\% } $ }\\

\hline
\hline
\end{tabular}
\end{table*}

\subsection{Comparison with full-reference quality assessment methods}

No-reference quality assessment is more challenging than full-reference methods, as it evaluates content without an original reference~\cite{saad2014blind, soundararajan2012video, qu2023lfacon}. This requires advanced algorithms to infer quality metrics directly from the content, simulating human perception without reference points~\cite{mittal2012making, mittal2015completely, qu2021light}. Consequently, developing no-reference methods demands a deeper understanding of perceptual quality and more sophisticated computational models~\cite{saad2014blind, madhusudana2022image, qu2023lfacon}.
Although NVS-SQA does not require references, we compared it with various full-reference quality assessment methods to determine if its performance is comparable to those that rely on references. Our extensive evaluation encompassed 16 widely recognized full-reference quality assessment methodologies, including PSNR, SSIM~\cite{wang2004image}, MS-SSIM~\cite{wang2003multiscale}, IW-SSIM~\cite{wang2010information}, VIF~\cite{sheikh2006image}, FSIM~\cite{zhang2011fsim}, GMSD~\cite{xue2013gradient}, VSI~\cite{zhang2014vsi}, DSS~\cite{balanov2015image}, HaarPSI~\cite{reisenhofer2018haar}, MDSI~\cite{nafchi2016mean}, LPIPS~\cite{zhang2018unreasonable}, DISTS~\cite{ding2020image}, STRRED~\cite{soundararajan2012video}, VMAF~\cite{li2016toward}, and FovVideoVDP~\cite{mantiuk2021fovvideovdp}.

Table~\ref{tab:fr_benchmarking} delineates the SRCC and PLCC metrics for these full-reference quality assessment methods alongside NVS-SQA, across the Fieldwork, LLFF, and Lab datasets. The table's concluding row highlights NVS-SQA's performance improvement over the second-highest scores. As noted in Table~\ref{tab:fr_benchmarking}, dataset LLFF lacks reference video data, making certain full-reference video-based methods such as STRRED, VMAF, and FovVideoVDP inapplicable. Overall, NVS-SQA achieves performance comparable to the evaluated full-reference methods and even demonstrates marginal improvements in some evaluation metrics. Specifically, in the LLFF dataset, NVS-SQA shows an enhancement with a 51.4\% increase in the SRCC and an 47.1\% rise in the KLCC relative to the top-performing full-reference methods. Moreover, NVS-SQA demonstrates the smallest standard deviation across most of the metrics and datasets, underscoring its stability compared to other methods. These results underscore the exceptional capability of the proposed no-reference method to outperform full-reference methods in the majority of evaluation scenarios, highlighting its potential.

\subsection{Ablation study}
\label{subsec:ablation_study}

\begin{table}[htb]
\centering
\caption{\textbf{Ablation study of individual branch contributions under manual (MBW) and auto-weighting (AQB) schemes.}
The upper block reports SRCC and PLCC for each single- and pairwise-branch combination using MBW. The lower block shows AQB performance when omitting each branch in turn (w/o IQA, w/o VQA, w/o REP) alongside the fully combined AQB (complete). The best score in each column is highlighted in bold.}

\label{tab:ablation}
\scriptsize
\renewcommand{\arraystretch}{1.2}
\setlength{\tabcolsep}{2pt}

\begin{tabular}{c|cc|cc|cc}
\hline
\hline
&\multicolumn{2}{c}{Fieldwork}&\multicolumn{2}{c}{LLFF}&\multicolumn{2}{c}{Lab}\\
\hline
Diff. Branches with MBW & SRCC ↑ & PLCC ↑ & SRCC ↑ & PLCC ↑ & SRCC ↑ & PLCC ↑ \\
\hline
IQA & $0.8556$ & $0.7873$ & $0.4625$ & $0.4433$ & $0.4571$ & $0.5183$\\
VQA & $0.6667$ & $0.5969$ & $0.5500$ & $0.4644$ & $0.6976$ & $0.6531$\\
IQA+VQA & $0.6778$ & $0.6848$ & $0.4875$ & $0.5480$ & $0.6874$ & $0.6643$\\
IQA+VQA+REP & $0.8778$ & $0.7956$ & $0.5875$ & $0.5512$ & $0.6429$ & $0.6503$\\
\hline
AQB w/o IQA       & $0.7832$ & $0.7439$ & $0.5827$ & $0.5041$ & $0.6514$ & $0.6422$ \\
AQB w/o VQA       & $0.8297$ & $0.8064$ & $0.6028$ & $0.5946$ & $0.5519$ & $0.5233$ \\
AQB w/o REP       & $0.8463$ & $0.8197$ & $0.6218$ & $0.6135$ & $0.6891$ & $0.6615$ \\
AQB (complete) & $\pmb{ 0.9111 }$ & $\pmb{ 0.8828 }$ & $\pmb{ 0.7000 }$ & $\pmb{ 0.6441 }$ & $\pmb{ 0.7000 }$ & $\pmb{ 0.6783 }$\\
\hline
\hline
\end{tabular}

\end{table}

\begin{figure}[htb]

\scriptsize
\renewcommand{\arraystretch}{1.2}
\setlength{\tabcolsep}{2pt}
\begin{tabular}{c|c} 
\hline\hline

\multicolumn{2}{c}{\textbf{Scene [Room] Synthesized by NVS methods (NeX) and (GNT-C)} }\\
\hline
\textit{Predicted Preference \textcolor{red}{w/o IQA Branch}} & \textit{Predicted Preference \textcolor{custom_green}{(with IQA Branch)}} \\
{\includegraphics[width=0.5\columnwidth, height=0.2\columnwidth]{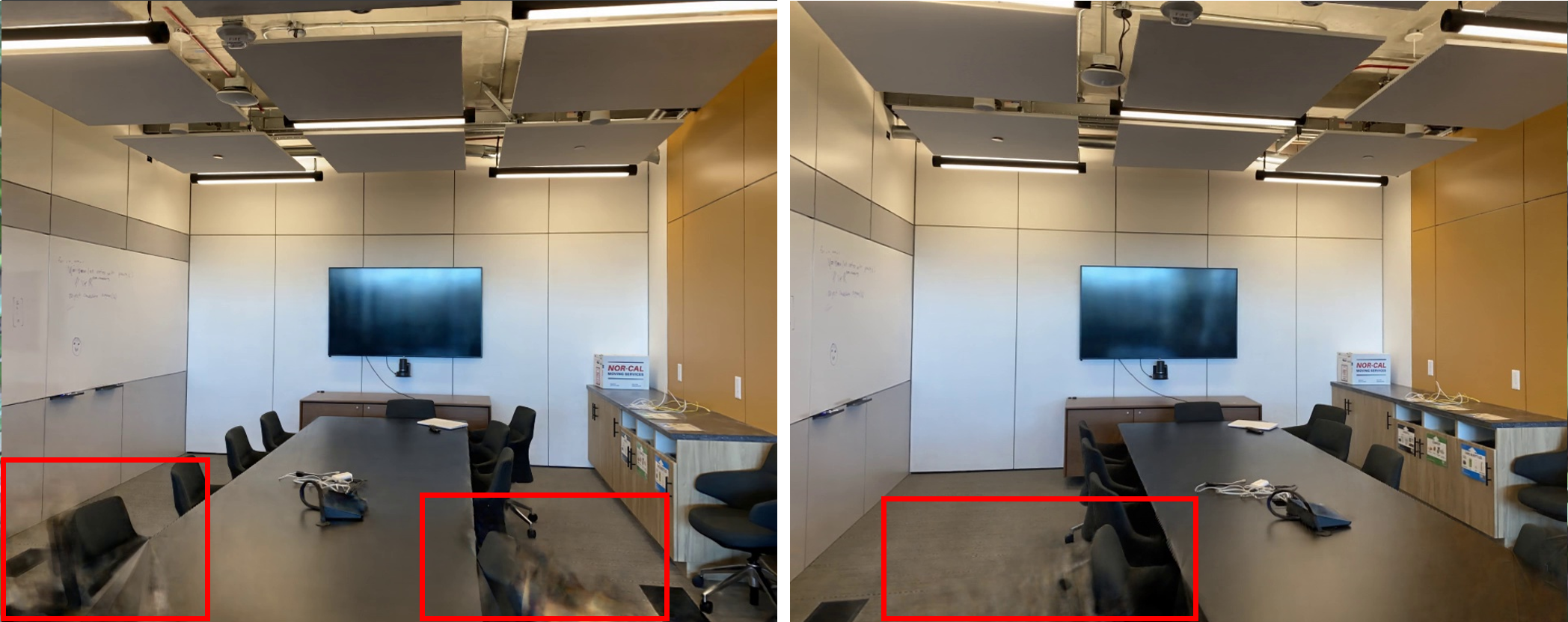}} & {\includegraphics[width=0.5\columnwidth, height=0.2\columnwidth]{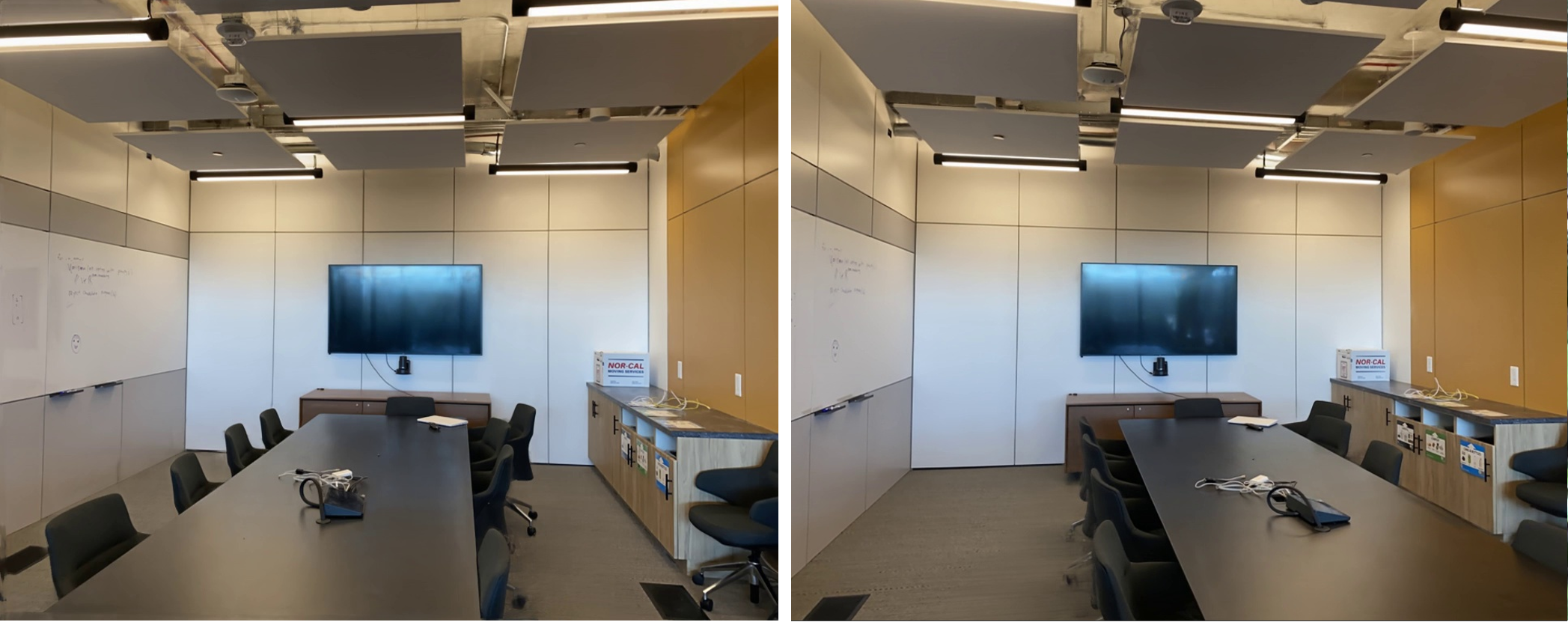}}\\


\hline\hline

\multicolumn{2}{c}{\textbf{Scene [Whale] Synthesized by NVS methods (GNT-C) and (IBRNet-S)} }\\
\hline
\textit{Predicted Preference \textcolor{red}{w/o VQA Branch}} & \textit{Predicted Preference \textcolor{custom_green}{(with VQA Branch)}} \\
{\includegraphics[width=0.5\columnwidth, height=0.2\columnwidth]{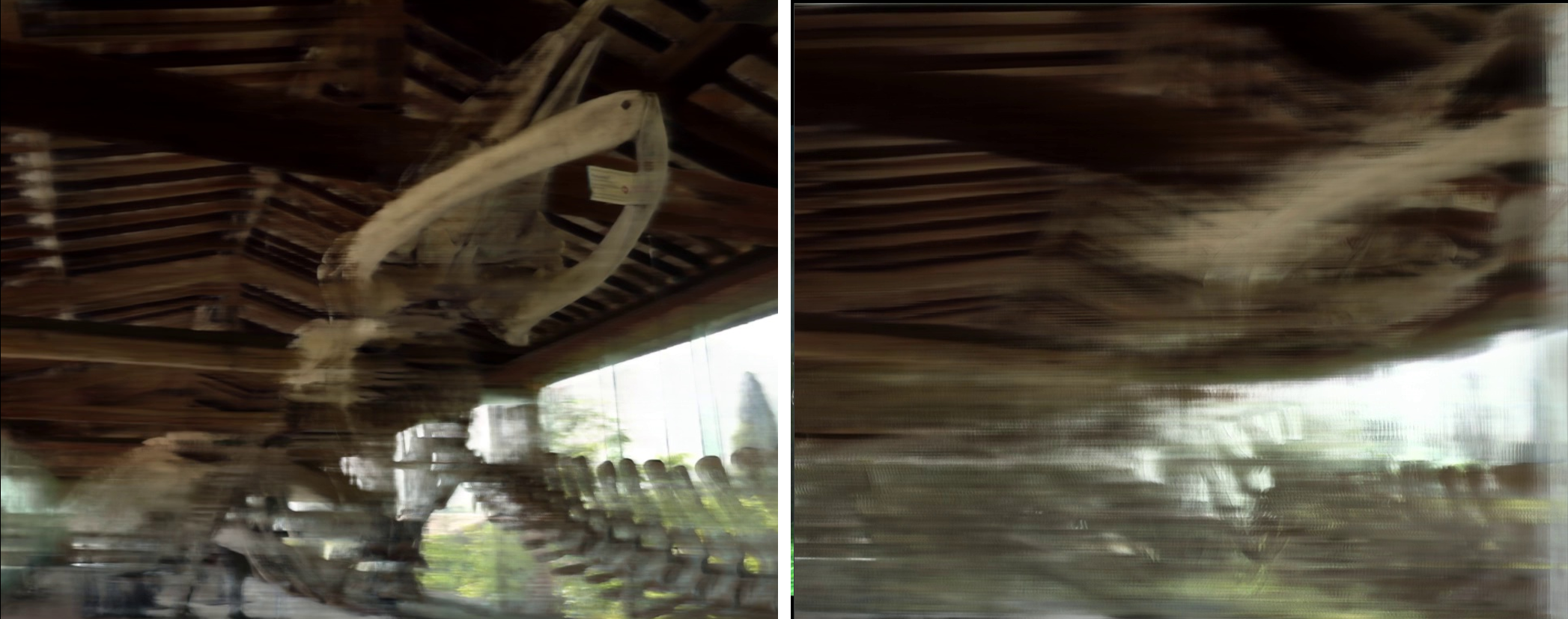}} & {\includegraphics[width=0.5\columnwidth, height=0.2\columnwidth]{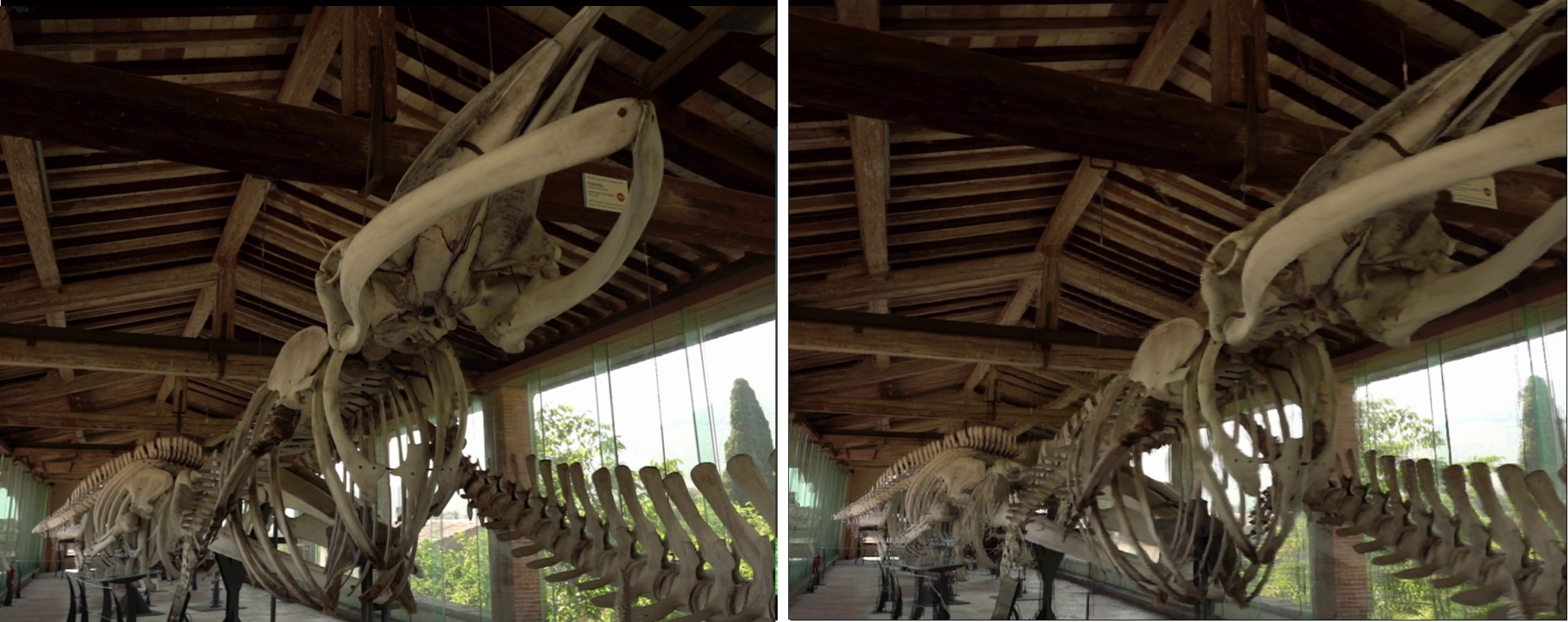}}\\


\hline\hline

\multicolumn{2}{c}{\textbf{Scene [Leaves] Synthesized by NVS methods (NeRF) and (NeX)} }\\
\hline
\textit{Predicted Preference \textcolor{red}{w/o REP Branch}} & \textit{Predicted Preference \textcolor{custom_green}{(with REP Branch)}} \\
{\includegraphics[width=0.5\columnwidth, height=0.2\columnwidth]{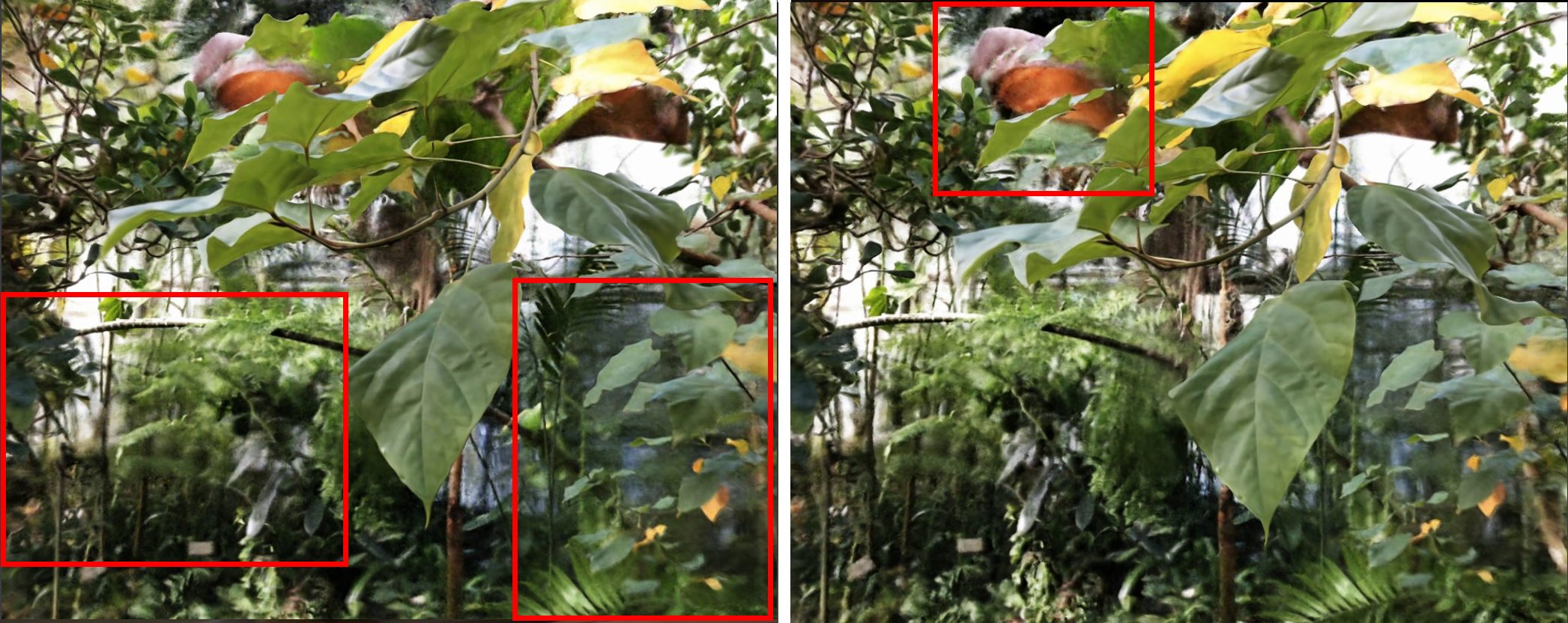}} & {\includegraphics[width=0.5\columnwidth, height=0.2\columnwidth]{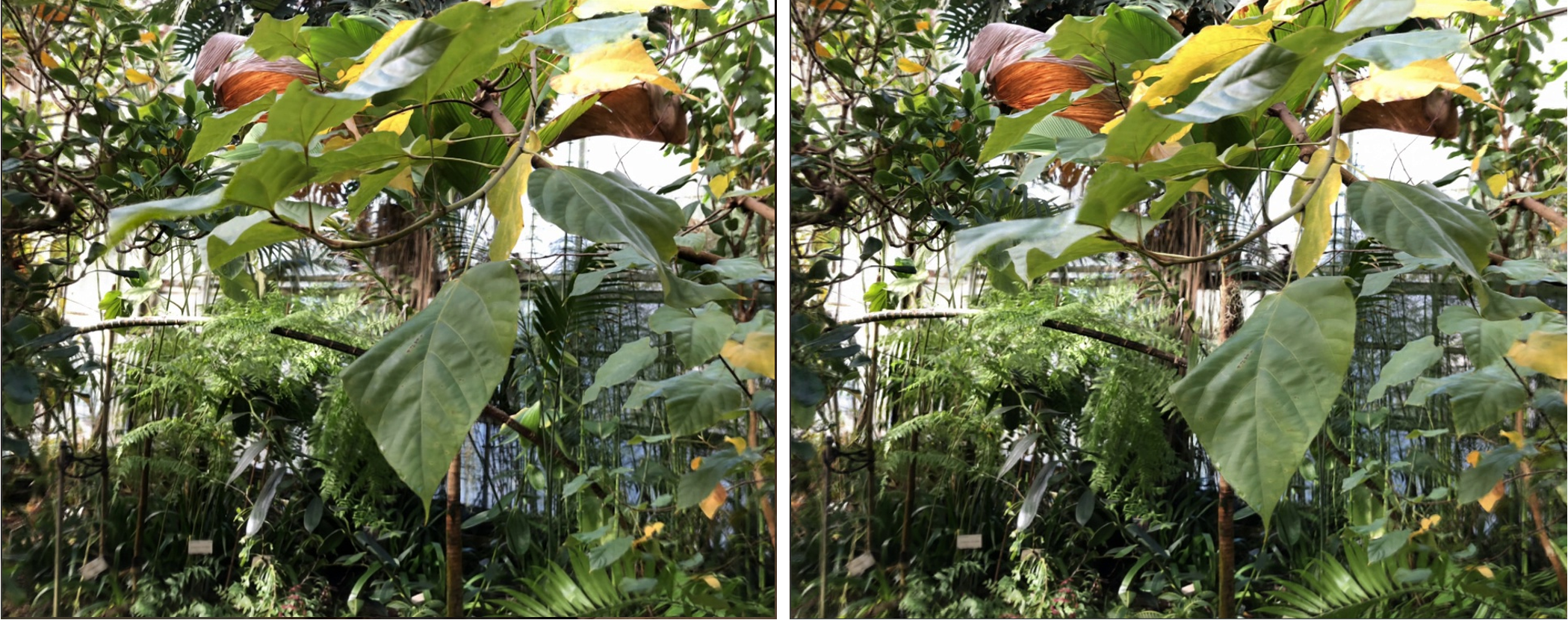}}\\


\hline\hline

\end{tabular}
\caption{\textbf{Qualitative illustration of the effects of individual guidance branches (IQA, VQA, REP) on the prediction preference of the proposed NVS-SQA model.} Each row shows two synthesized scenes generated by different NVS methods. The left and right columns indicate which scene is preferred (i.e., predicted as higher quality) by NVS-SQA when trained without or with a specific guidance branch. For example, when the IQA branch is omitted, the model tends to favor scenes with blurred or texture-inconsistent regions, while inclusion of the IQA branch shifts preference toward perceptually sharper results. Similar trends are observed for the VQA and REP branches, where their inclusion improves sensitivity to temporal consistency and structural integrity, respectively. These results demonstrate that each guidance branch contributes a distinct and complementary perceptual cue to the overall quality representation (zooming in for a clearer view).}
\label{fig:design_choise}
\end{figure}

We first apply manual weighting MBW (as described in Section~\ref{subsubsec:mbw}) with a comprehensive grid search to ascertain the optimal combination of weights for the learning objectives, as specified in Equation~\ref{eq:obj}, within the search space $\{ 0, \; 0.1,\; 0.2,\; 0.5,\; 1,\; 1.5,\; 2 \}$. The search culminated in the identification of the most effective weight combination: $\{ \lambda_{IQA}: 1.5, \;\lambda_{VQA}: 1, \; \lambda_{REP}: 0.2  \}$. We then applied the auto-weighting method AQB (introduced in Section~\ref{subsubsec:aqb}) to evaluate its effectiveness. Table~\ref{tab:ablation} presents the results of ablation studies, assessing the individual and collective contributions of each learning branch, as well as the impact of AQB. The studies conducted on Fieldwork, LLFF, and Lab demonstrate that combining all branches in MBW outperforms using fewer branches, with AQB further enhancing performance. Moreover, we conduct an extended ablation study to analyze the contribution of each guidance branch (IQA, VQA, and REP) under both manual weighting (MBW) and the proposed adaptive scheme (AQB). As shown in Table~\ref{tab:ablation}, we report results for each individual branch, their combinations, and the complete model. In addition, we include AQB-based ablations that selectively remove one branch at a time. These results show that the IQA branch is critical, with a marked performance degradation observed upon its removal. The VQA branch also plays a significant role, particularly in capturing angular and temporal coherence. Meanwhile, the REP branch provides complementary quality features not captured by IQA or VQA, further enhancing overall performance. This comprehensive analysis further validates the importance of each component in our multi-branch design and the effectiveness of AQB in leveraging them optimally. Fig.~\ref{fig:design_choise} qualitatively illustrates the influence of each guidance branch on the model’s prediction preference. When any of the three branches is removed, NVS-SQA tends to assign higher quality scores to perceptually inferior scenes, such as those exhibiting blurring, motion instability, or structural misalignment, revealing a reduced sensitivity to the corresponding perceptual dimension.

To evaluate the sensitivity of NVS-SQA to training hyperparameters, we varied the number of training epochs and batch sizes while keeping all other factors constant, including the architecture, AQB loss function, and unlabeled training data. The results, presented in Table~\ref{tab:hyperparam_sensitivity}, indicate that NVS-SQA maintains robust performance across a range of batch sizes when the number of training epochs is sufficient. This suggests that the model is not overly sensitive to moderate changes in these hyperparameters.

\begin{table}[htb]
\centering
\caption{\textbf{Sensitivity analysis of training hyperparameters.} This table reports the SRCC, PLCC, and KRCC values under different combinations of epochs and batch sizes using the Fieldwork dataset.}
\label{tab:hyperparam_sensitivity}
\footnotesize
\renewcommand{\arraystretch}{1.2}
\setlength{\tabcolsep}{8pt}

\begin{tabular}{c|c|S[table-format=1.4] S[table-format=1.4] S[table-format=1.4]}
\hline
\hline
Epochs & Batch Size & {SRCC $\uparrow$} & {PLCC $\uparrow$} & {KRCC $\uparrow$} \\
\hline
100            & 16 & 0.8352 & 0.7823 & 0.7538 \\
200            & 16 & \pmb{0.9110} & \pmb{0.8831} & \pmb{0.8226} \\
300            & 16 & 0.9034 & 0.8707 & 0.8142 \\
200            & 8  & 0.8796 & 0.8549 & 0.7931 \\
200            & 32 & 0.9012 & 0.8684 & 0.8107 \\
\hline
\hline
\end{tabular}

\end{table}

To evaluate the contribution of our proposed contrastive pair preparation, we introduced a baseline that removes the entire pairing mechanism. Specifically, this variant samples NSS pairs randomly across the dataset without ensuring semantic consistency, view replacement, or controlled distortion. The results can be found in the first and last row of Table~\ref{tab:ablation_cpp_architecture}. The absence of these constraints substantially degrades the model’s ability to learn meaningful quality representations: for example, SRCC on the Fieldwork dataset drops from 0.911 to 0.745. This confirms that our contrastive pair preparation strategy provides essential structure for learning discriminative quality embeddings.

In addition, we explored several alternatives to our Transformer-based angular fusion design within AdaptiSceneNet. We replaced the Transformer module with convolution-only layers, GRUs, and LSTMs, respectively. The results can be found in Table~\ref{tab:ablation_cpp_architecture}. While both GRU and LSTM architectures offered improvements over CNN-only designs, they consistently underperformed compared to our Transformer-based approach. For example, on the Fieldwork dataset, SRCC improved from 0.714 (CNN-only) to 0.820 (CNN+LSTM), but remained below the 0.911 achieved by our full model. These results validate the advantage of using self-attention to model long-range dependencies across views, which is critical for capturing angular consistency in NSS.

\begin{table}[htb]
\centering
\caption{\textbf{Ablation on contrastive pair preparation and angular fusion design.} This table reports SRCC values on the Fieldwork, LLFF, and Lab datasets using different architectural and training configurations.}
\label{tab:ablation_cpp_architecture}
\scriptsize
\renewcommand{\arraystretch}{1.2}
\setlength{\tabcolsep}{8pt}

\begin{tabular}{l|S[table-format=1.4] S[table-format=1.4] S[table-format=1.4]}
\hline
\hline
\textbf{Model Variant} & {Fieldwork $\uparrow$} & {LLFF $\uparrow$} & {Lab $\uparrow$} \\
\hline
w/o Contrastive Pair Preparation & 0.7452 & 0.6651 & 0.6874 \\
CNN-only                         & 0.7147 & 0.6303 & 0.6182 \\
CNN + GRU                        & 0.7856 & 0.7189 & 0.7130 \\
CNN + LSTM                       & 0.8203 & 0.7458 & 0.7507 \\
CNN + Transformer (Ours)         & \textbf{0.9110} & \textbf{0.8831} & \textbf{0.8226} \\
\hline
\hline
\end{tabular}

\end{table}

\begin{table}[htb]
\centering
\caption{\textbf{Comparison with the fully-supervised NeRF-NQA.} using SRCC and PLCC metrics across the Fieldwork, LLFF, and Lab datasets. Optimal results in each column are highlighted in bold. Given NeRF-NQA was end-to-end trained on perceptual labels, NVS-SQA is also fine-tuned to ensure a fair comparison.}
\label{tab:compare_nerf_nqa}
\footnotesize
\renewcommand{\arraystretch}{1.2}
\setlength{\tabcolsep}{2pt}
\begin{tabular}{c|cc|cc|cc}
\hline
\hline
&\multicolumn{2}{c}{Fieldwork}&\multicolumn{2}{c}{LLFF}&\multicolumn{2}{c}{Lab}\\
\hline
Method & SRCC ↑ & PLCC ↑ & SRCC ↑ & PLCC ↑ & SRCC ↑ & PLCC ↑ \\
\hline

NeRF-NQA & $0.8667$ & $0.8452$ & $0.7850$ & $0.7614$ & $0.8145$ & $0.7774$\\
NVS-SQA & $\pmb{ 0.9389 }$ & $\pmb{ 0.9487}$ & $\pmb{ 0.8580 }$ & $\pmb{ 0.8346 }$ & $\pmb{ 0.8863 }$ & $\pmb{ 0.8660 }$\\

\hline
\hline
\end{tabular}
\end{table}

\subsection{Comparison with the fully supervised method.}

Furthermore, we conducted a comparative analysis between the proposed NVS-SQA framework and our previously introduced fully-supervised method, NeRF-NQA~\cite{qu2024nerfnqa}. Given that NeRF-NQA was developed utilizing perceptual labels for training, we accordingly fine-tuned NVS-SQA to ensure an equitable comparison. As delineated in Table~\ref{tab:compare_nerf_nqa}, our method surpasses NeRF-NQA across all evaluation metrics within the three datasets under consideration. Beyond mere performance metrics, NVS-SQA distinguishes itself as an end-to-end learning framework that facilitates straightforward training, in contrast to NeRF-NQA, which does not support end-to-end learning~\cite{qu2024nerfnqa}. Specifically, end-to-end learning facilitates the direct transformation of raw data into final outputs, enhancing model efficiency and accuracy by eliminating manual feature engineering~\cite{lecun2015deep, he2016deep}. This approach not only simplifies the development process but also improves the scalability and adaptability~\cite{lecun2015deep}. Additionally, unlike NeRF-NQA, which relies on external tools such as COLMAP~\cite{schonberger2016structure}—known for generating noisy sparse points in some cases~\cite{darmon2022improving, bai2024colmap}—NVS-SQA operates independently. This independence further emphasizes its practicality and reliability for the quality assessment of NSS, as it does not depend on external systems that may introduce noise or errors into the process.

\subsection{Quantitative results}

In addition to the numerical comparisons reported in Table X, we provide example cases in Fig.~\ref{fig:example_scenes}, where PSNR, SSIM, and LPIPS (i.e., the most widely used quality assessment methods for NSS) fail to align with actual human preferences. Each row in Fig.~\ref{fig:example_scenes} presents a pair of neurally synthesized scenes (NSS) generated by different Neural View Synthesis (NVS) methods, along with human and NVS-SQA preferences, compared against the outcomes from PSNR, SSIM, and LPIPS. For instance, in the top row (“Giraffe”), although PSNR and SSIM favor the right-hand result, perceptual inspection reveals substantial blurring and missing object details. By contrast, NVS-SQA and human observers correctly identify the left-hand image as exhibiting superior perceptual fidelity. A similar discrepancy is evident in the “Intr-Animals” example, where the left-hand NSS contains clear artifacts (outlined in red boxes) that standard metrics overlook, yet NVS-SQA accurately detects. In the “Orchids” scene, inconsistent color hues across views go largely unnoticed by PSNR, SSIM, or LPIPS but are recognized by our proposed method as detrimental to perceptual quality.

\begin{table}[htb]
\centering
\caption{\textbf{Comparison of model complexity and inference time.} The table shows the number of parameters and the average inference time per NSS for representative quality assessment methods from different categories.}
\label{tab:params_inference}
\footnotesize
\renewcommand{\arraystretch}{1.2}
\setlength{\tabcolsep}{6pt}

\begin{tabular}{c|l|S[table-format=2.1] | S[table-format=2.3]}
\hline
\hline
Type & Method & {Params. (M) ↓} & {Proc. Time per NSS (s) ↓} \\
\hline
\multirow{2}{*}{IQA}   & CONTRIQUE & 30.1 & 34.168 \\
                       & Re-IQA    & 68.5 & 57.436 \\
\hline
\multirow{2}{*}{VQA}   & FAST-VQA  & 27.7 &  \phantom{1} \underline{0.743} \\
                       & DOVER     & 55.5 &  0.801 \\
\hline
\multirow{2}{*}{LFIQA} & ALAS-DADS &  \phantom{1} \pmb{1.5} &  1.306 \\
                       & LFACon    & 10.5 &  0.979 \\
\hline
\multicolumn{2}{c|}{\textbf{NVS-SQA (Ours)}}     &  \phantom{1} \underline{4.7} &  \phantom{1} \pmb{0.692} \\
\hline
\hline
\end{tabular}

\end{table}

\begin{figure}[htb]

\scriptsize
\renewcommand{\arraystretch}{1.2}
\setlength{\tabcolsep}{2pt}
\begin{tabular}{c|c} 
\hline\hline

\multicolumn{2}{c}{\textbf{Scene [Giraffe] Synthesized by NVS methods (NeRF) and (IBRNet-C)} }\\
\hline
{\includegraphics[width=0.5\columnwidth, height=0.2\columnwidth]{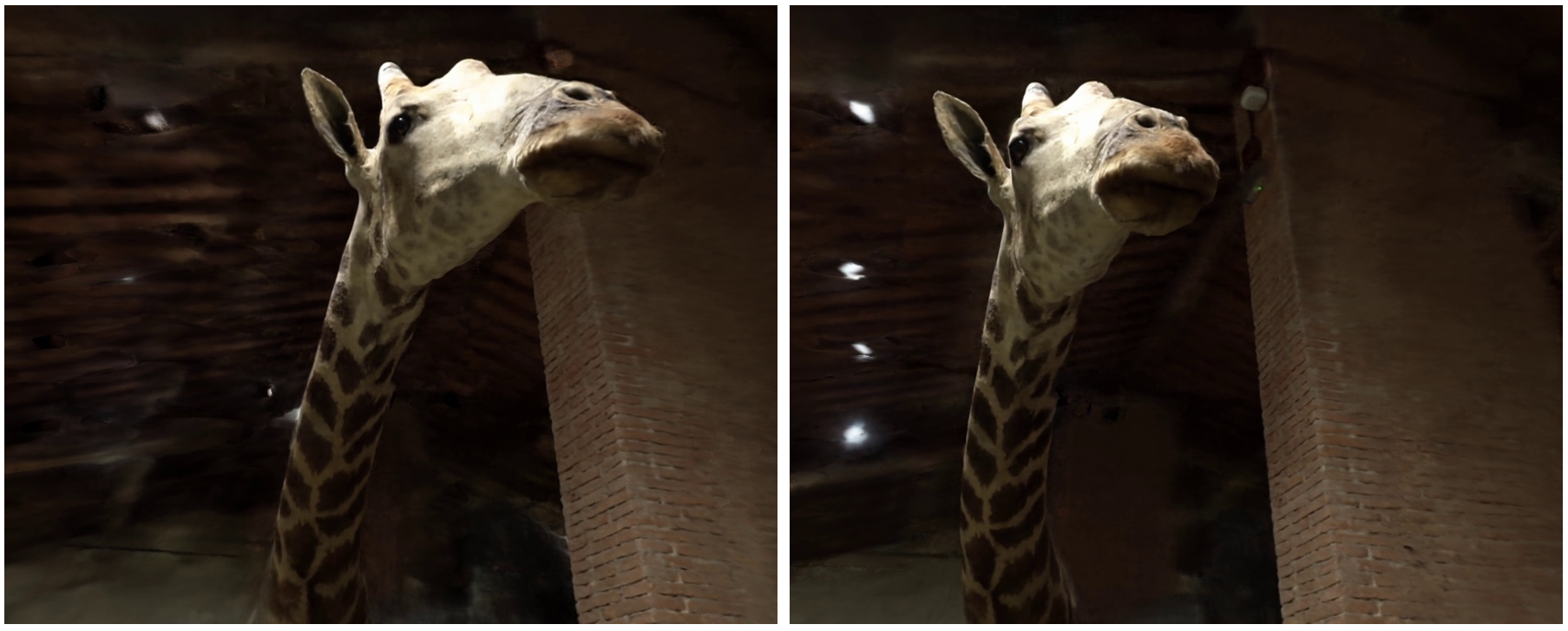}} & {\includegraphics[width=0.5\columnwidth, height=0.2\columnwidth]{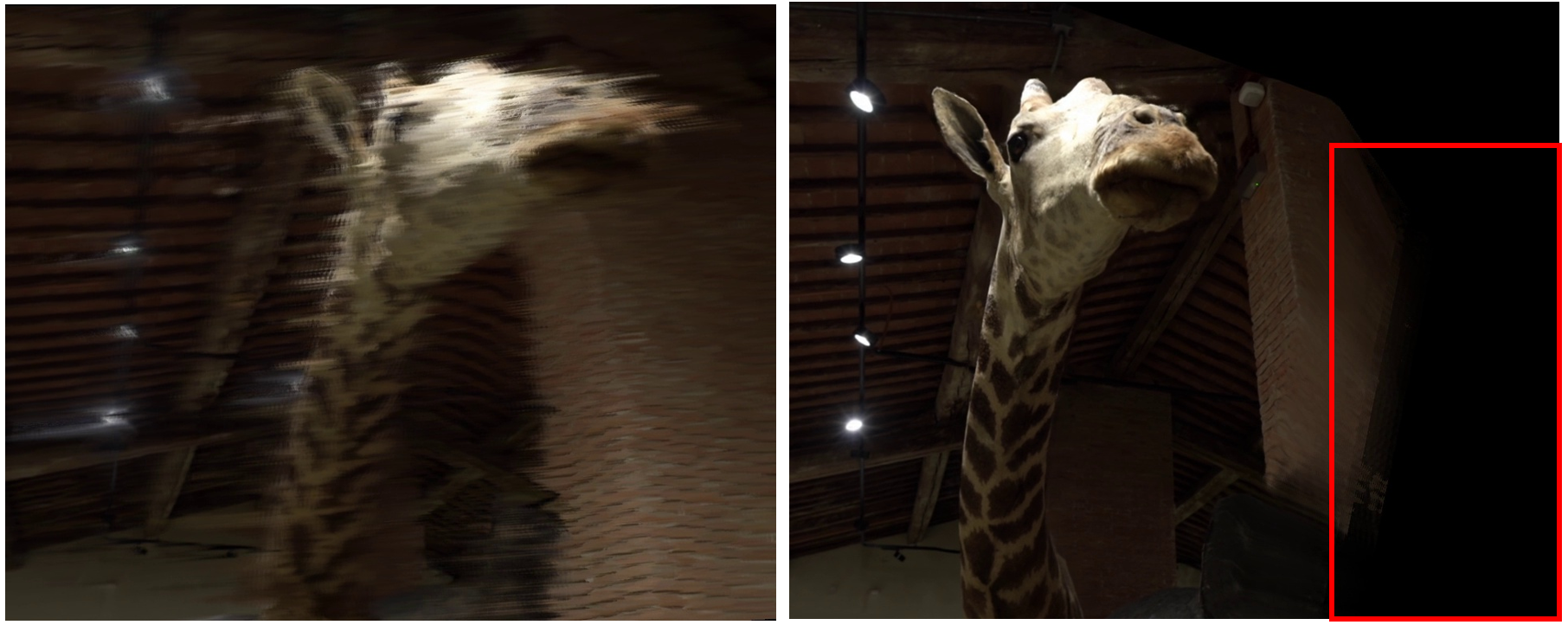}}\\
\hline
\textit{\textcolor{custom_green}{Human, and NVS-SQA Preferences}} & \textit{\textcolor{red}{PSNR, SSIM, and LPIPS Preferences}} \\
\hline
\multicolumn{2}{c}{\textbf{Reason}: The right NSS exhibits noticeable blurring and missing content.
}\\

\hline\hline

\multicolumn{2}{c}{\textbf{Scene [Intr-Animals] Synthesized by NVS methods (DVGO) and (LFNR)} }\\
\hline
{\includegraphics[width=0.5\columnwidth, height=0.2\columnwidth]{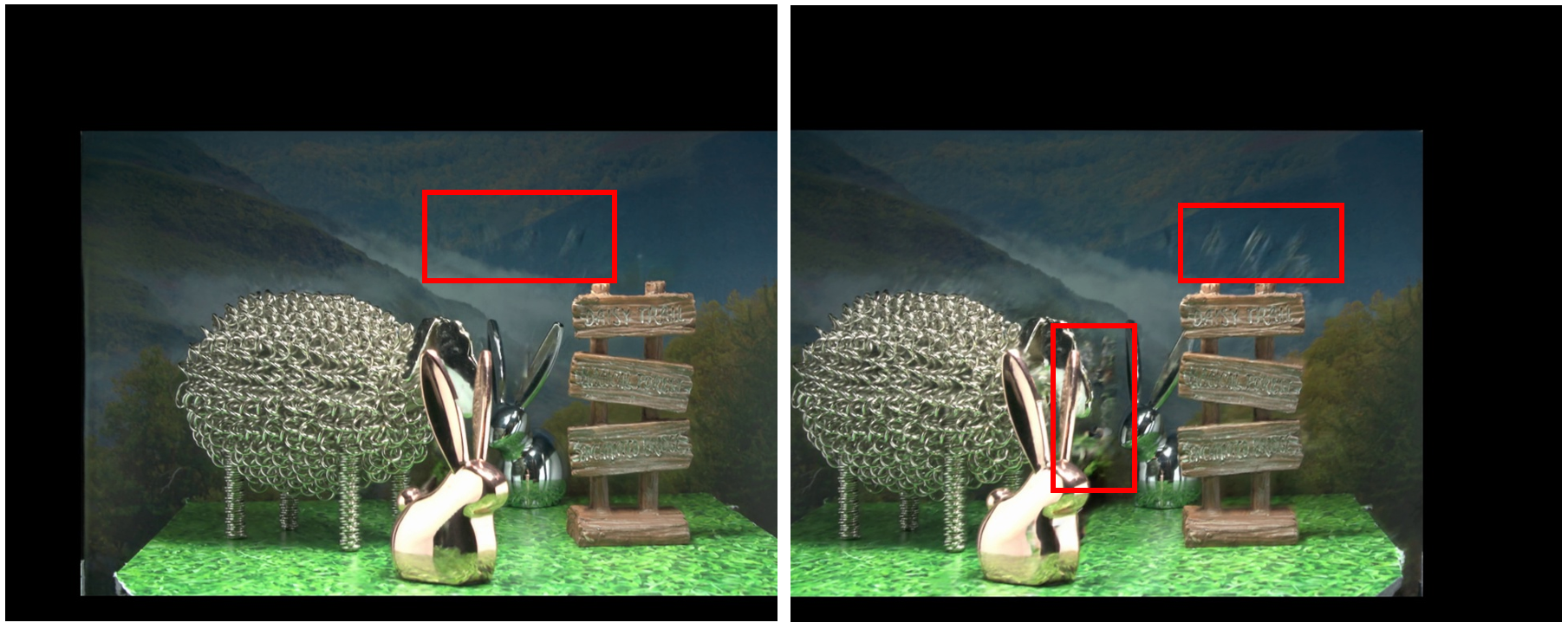}} & {\includegraphics[width=0.5\columnwidth, height=0.2\columnwidth]{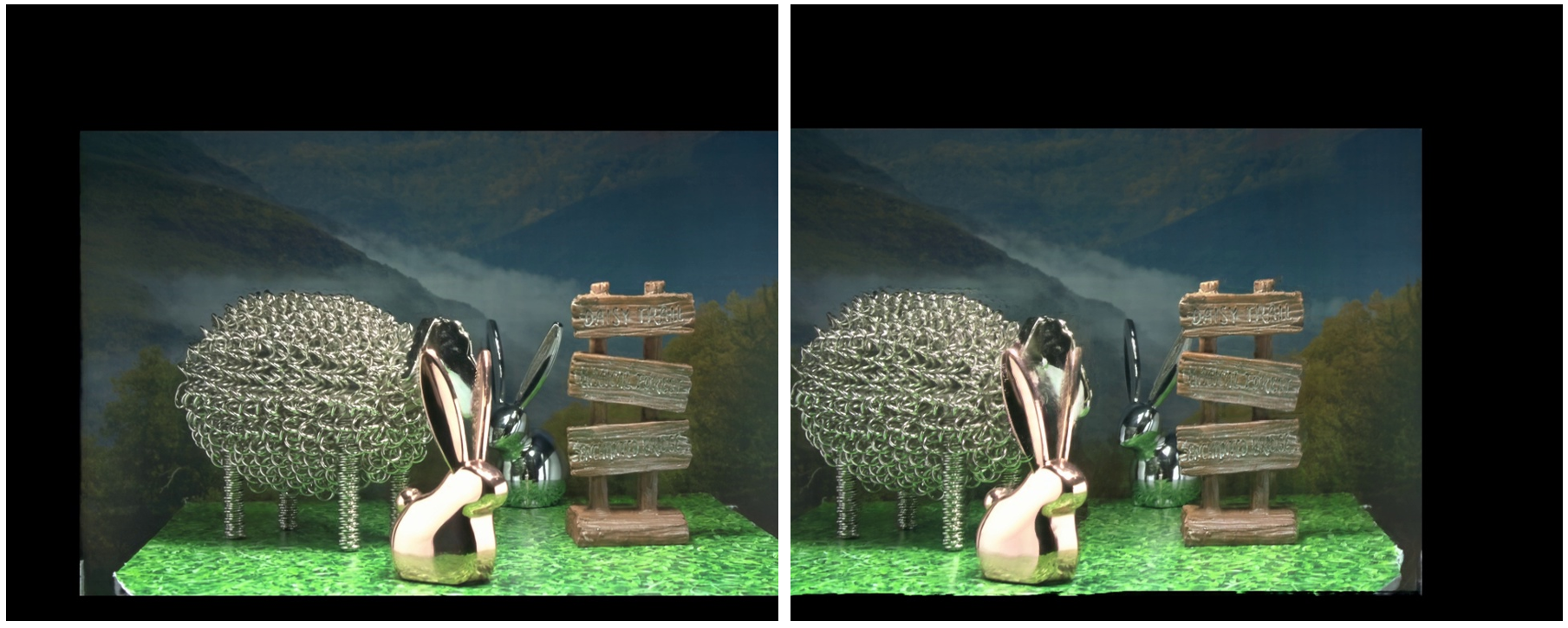}}\\
\textit{\textcolor{red}{PSNR, SSIM, and LPIPS Preferences}} & \textit{\textcolor{custom_green}{Human, and NVS-SQA Preferences}} \\
\hline
\multicolumn{2}{c}{\textbf{Reason}: The left NSS shows clear artifacts highlighted by red boxes.
}\\

\hline\hline

\multicolumn{2}{c}{\textbf{Scene [Orchids] Synthesized by NVS methods (GNT-C) and (NeX)} }\\
\hline
{\includegraphics[width=0.5\columnwidth, height=0.2\columnwidth]{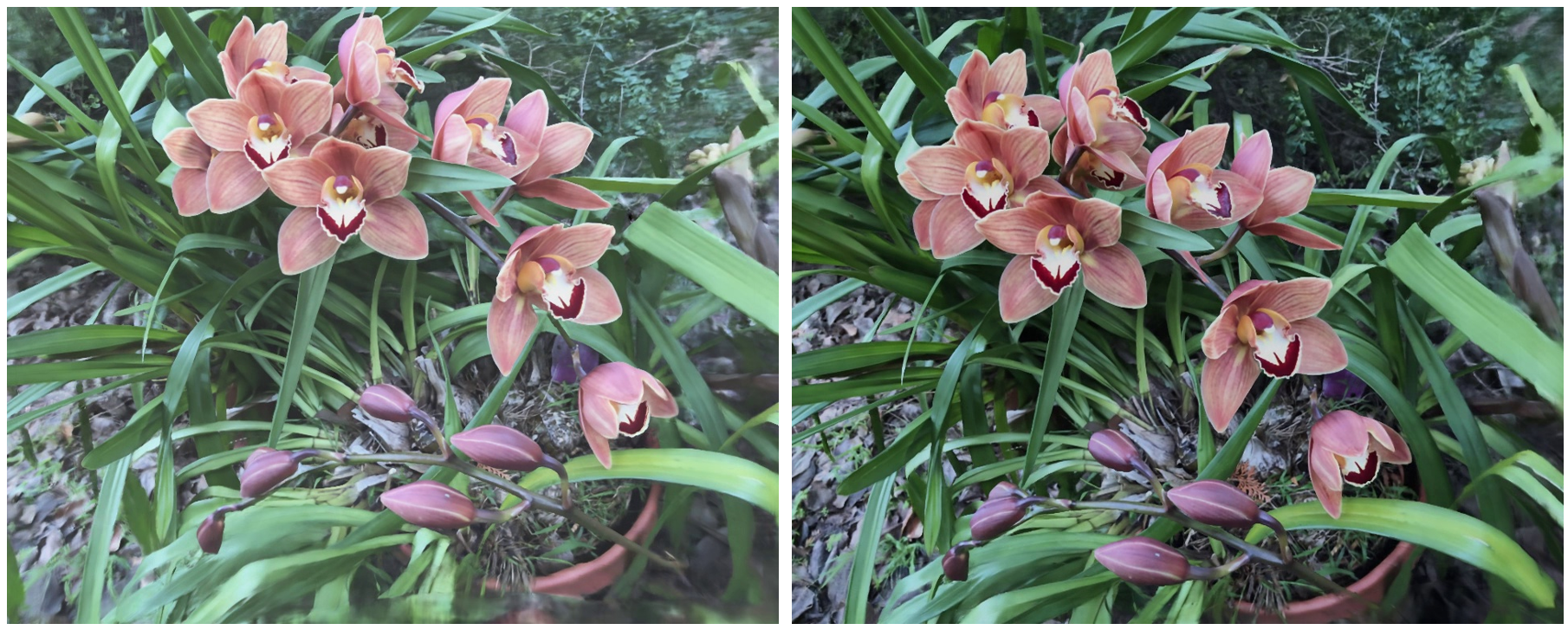}} & {\includegraphics[width=0.5\columnwidth, height=0.2\columnwidth]{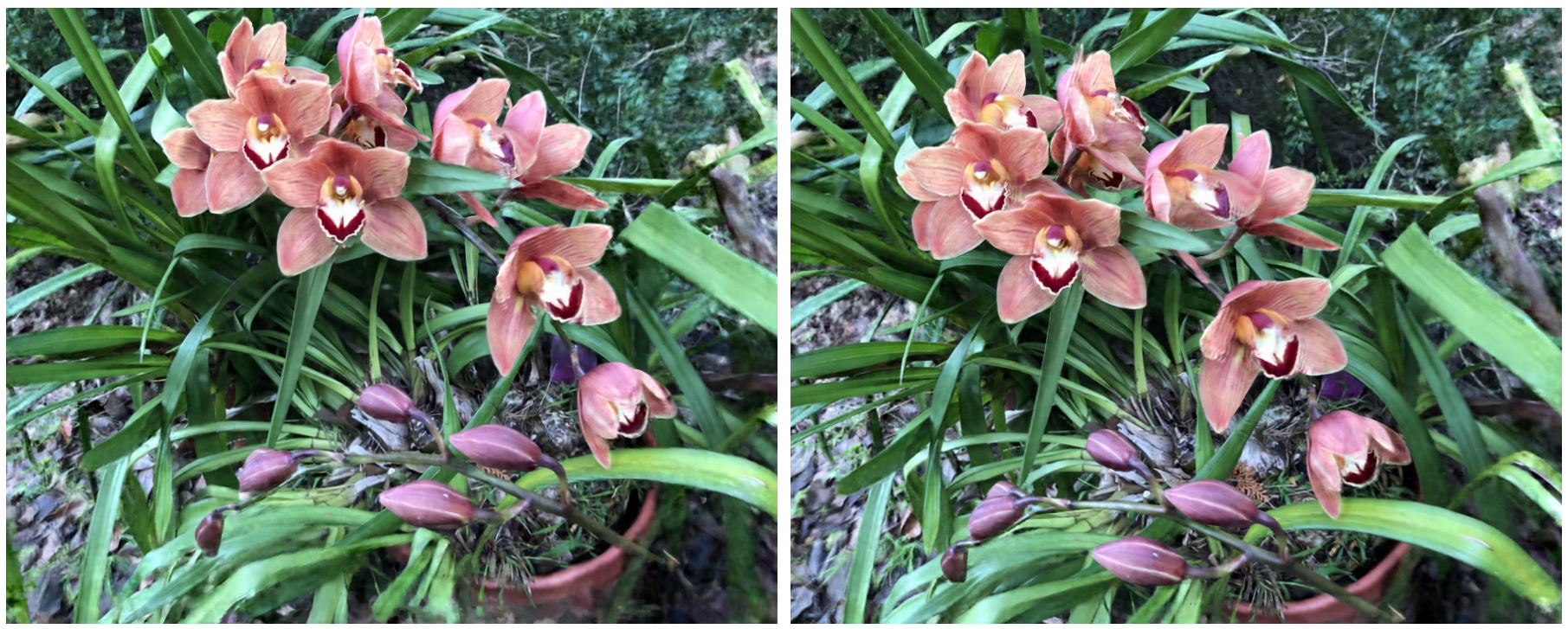}}\\
\textit{\textcolor{red}{PSNR, SSIM, and LPIPS Preferences}} & \textit{\textcolor{custom_green}{Human, and NVS-SQA Preferences}} \\
\hline
\multicolumn{2}{c}{\textbf{Reason}: The left NSS displays inconsistent color hues across views.
}\\

\hline\hline

\end{tabular}
\caption{\textbf{Example NSS generated by various NVS methods.} In all cases, the proposed NeRF-NQA aligns with human preferences without references, whereas the prevalent full-reference methods PSNR, SSIM, and LPIPS fall short (zoom in for a clearer view).}
\label{fig:example_scenes}
\end{figure}

\subsection{Computational Cost Analysis}
\label{subsec:computational_cost}

We present a detailed analysis of the computational efficiency and model complexity of NVS-SQA in comparison with representative no-reference IQA, VQA, and LFIQA methods. As summarized in Table~\ref{tab:params_inference}, NVS-SQA demonstrates the lowest inference latency, requiring only 0.692 seconds per NSS on average. In addition, it maintains a relatively compact architecture with only 4.7 million parameters, making it significantly more efficient than most existing methods in both runtime and model size.

\subsection{Robustness Evaluation in Extreme Scenarios}
\label{subsec:robustness_test}
To further examine the generalization capability of the proposed NVS-SQA model in practical conditions, we conducted a dedicated robustness evaluation under extreme scenarios that are typically challenging for novel view synthesis. These scenarios include highly sparse input views, high-noise environments, and complex lighting conditions. Such cases frequently occur in real-world applications but are often underrepresented in conventional evaluation datasets.

\noindent{\textbf{Evaluation setup and human label collection.}}
We selected five challenging scenes (namely scan8, scan55, scan103, scan110, and scan114) from the DTU dataset~\cite{jensen2014large}, characterized by extreme conditions such as highly sparse input views, significant noise levels, and complex lighting environments. For each scene, five neural view synthesis methods were used to generate the corresponding NSS: 2DGS~\cite{huang20242d}, 3D Gaussian Splatting~\cite{kerbl20233d}, 3DLS~\cite{chen2024beyond}, AbsGS~\cite{ye2024absgs}, and Mip-Splatting~\cite{yu2024mip}. This resulted in a total of 25 synthesized scenes across all test cases. To obtain reliable perceptual quality labels for these NSS, we conducted a user study of pairwise comparison~\cite{handley2001comparative, paudyal2017towards, zhang2018unreasonable} involving 31 participants (mean age = 25.13, SD = 1.2; 18 males, 13 females). Each participant was asked to compare ten method pairs within each scene (i.e., all unique combinations among the five methods), resulting in 50 pairwise comparisons per participant. The order of presentation and spatial placement of video sequences were randomized to eliminate bias. The collected pairwise preference data were then converted into scalar quality scores using the Bradley-Terry model~\cite{hunter2004mm}, a statistical framework widely used for estimating latent scores from comparative data. This approach enabled the derivation of perceptual scores for each synthesized scene without requiring any reference views.

\noindent{\textbf{Results and Analysis.}}
The same model weights from previous experiments were used, without any fine-tuning or adaptation. As shown in Table~\ref{tab:robustness_extreme}, NVS-SQA achieved a SRCC of 0.6000, a PLCC of 0.7707, and a KRCC of 0.5333. Although all evaluated methods exhibit performance degradation in extreme scenarios, the proposed method still considerably outperforms existing quality assessment methods. For example, the second-best method, FAST-VQA, achieved an SRCC of only 0.3000 and a PLCC of 0.3614. These results indicate that NVS-SQA maintains perceptual alignment with human judgments even under adverse visual conditions, effectively capturing degradations associated with sparse geometry, noise-induced artifacts, and illumination inconsistencies, without requiring any reference views or domain-specific adjustment.

\begin{table}[htb]
\centering
\caption{\textbf{Robustness Evaluation under Extreme Scenarios.} Performance of representative quality assessment methods on neurally synthesized scenes with sparse views, noise, and complex lighting conditions. NVS-SQA shows the highest consistency with human perception.}
\label{tab:robustness_extreme}
\footnotesize
\renewcommand{\arraystretch}{1.2}
\setlength{\tabcolsep}{6pt}

\begin{tabular}{c|l|S[table-format=1.4] S[table-format=1.4] S[table-format=1.4]}
\hline
\hline
Type & Method & {SRCC ↑} & {PLCC ↑} & {KRCC ↑} \\
\hline
\multirow{6}{*}{IQA} 
& TV           & -0.4732 & -0.6783 & -0.4300 \\
& BRISQUE      &  0.0268 &  0.2121 & -0.0300 \\
& NIQE         & -0.1000 &  0.0293 & -0.1333 \\
& CLIP-IQA     & -0.3732 & -0.5991 & -0.2966 \\
& CONTRIQUE    &  0.1268 &  0.0239 &  0.1034 \\
& Re-IQA       & -0.1732 & -0.2981 & -0.1633 \\
\hline
\multirow{3}{*}{VQA} 
& VIIDEO       &  0.1000 &  0.2550 &  0.1333 \\
& FAST-VQA     &  0.3000 &  0.3614 &  0.2667 \\
& DOVER & 0.2732 & 0.3136 & 0.2966 \\
\hline
\multirow{2}{*}{LFIQA} 
& ALAS-DADS    &  0.1217 &  0.1837 &  0.1395 \\
& LFACon       &  0.2643 &  0.3215 &  0.2197 \\
\hline
\multicolumn{2}{c|}{\textbf{NVS-SQA (Ours)}} & \pmb{0.6000} & \pmb{0.7707} & \pmb{0.5333} \\
\hline
\hline
\end{tabular}

\end{table}

\section{Limitations and Future Work}

While our extensive evaluations (Section~\ref{subsec:robustness_test}) demonstrate that NVS-SQA remains relatively stable under challenging conditions such as resolution variations and severe occlusions, it may still be limited in handling more complex motion dynamics and more diverse scene types. On the one hand, although the proposed method maintains strong correlations with human perceptual judgments, subtle degradations in scenes involving complex motion dynamics, such as rapid movements, dynamic illumination changes, or intricate texture transitions, may not always be fully captured. On the other hand, our current self-supervised training pipeline is primarily constructed using realistic 3D reconstructed scenes, which rarely contain severe synthesis artifacts found in synthetic scenes, such as large black holes, view extrapolation failures, or strong shadow distortions. This limitation may reduce the model’s sensitivity to these rare yet perceptually critical defects in synthetic scenes. In future work, we plan to enrich the training data by incorporating more diverse NVS sources and introducing controlled, manually distorted samples, particularly those exhibiting severe occlusions and shadow artifacts, to enhance the robustness and generalizability of the learned quality representations across more complex motion dynamics and diverse scene types. Another potential limitation may lie in the use of the current guidance cues, which may not yet fully exploit the complementary strengths inherent in NVS-SQA. While the proposed guidance cues (IQA, VQA, and REP) are grounded in well-established perceptual characteristics of neural scene synthesis and provide effective, complementary supervision, there may remain subtle aspects of perceptual quality that are not fully captured by these signals. Future work could investigate more flexible or data-driven supervision strategies, such as leveraging learned perceptual priors or incorporating adversarial feedback mechanisms, to further enhance the representational alignment with nuanced human perception.

\section{Conclusion}

We present NVS-SQA, the first no-reference, self-supervised learning framework for NSS quality assessment, addressing challenges with unlabeled and limited datasets. By introducing NSS-specific contrastive pair preparation and multi-branch guidance adaptation inspired by heuristic cues and full-reference scores, NVS-SQA surpasses existing no-reference methods and even several full-reference metrics across diverse datasets, demonstrating strong generalization to unseen scenes and NVS methods. Additionally, we establish a benchmark for self-supervised NSS quality assessment and open-source our code and datasets to advance research in this domain, setting a new standard for NSS quality learning.

\bibliographystyle{IEEEtran}
\bibliography{references}





\end{document}